%% file: main.tex
\documentclass[11pt, a4paper,logo,  copyright, nonumbering]{paddleocr}
\usepackage[numbers]{natbib}
\usepackage{dblfloatfix}
\usepackage{ulem}
\usepackage{float}
\usepackage{tocloft}
\usepackage{caption}
\usepackage{dramatist}
\usepackage{xspace}
\usepackage{array}
\usepackage{pifont} %
\usepackage{multirow}
\usepackage{geometry}
\usepackage{makecell}
\usepackage{tcolorbox}
\usepackage{xltabular}
\usepackage{titlesec} 
\usepackage{longtable}
\usepackage[hang, flushmargin]{footmisc}
\usepackage{booktabs} 
\usepackage{xcolor}   
\usepackage[table]{xcolor}
\definecolor{lightpink}{RGB}{255, 182, 193}
\definecolor{iccvblue}{rgb}{0.21,0.49,0.74}

\hypersetup{
    colorlinks=true,
    linkcolor=blue,
    citecolor=blue,
    filecolor=blue,
    urlcolor=blue,
    pdfpagemode=UseNone,
    pdfstartview=FitH
}
\usepackage{threeparttable} 
\usepackage{soul}
\usepackage{amsfonts}
\usepackage{amsmath}
\usepackage{amssymb}
\usepackage{lineno}
\usepackage{multirow}
\usepackage{adjustbox}

\usepackage{CJKutf8}
\usepackage{subcaption}
\usepackage{setspace}

\usepackage{dsfont}
\usepackage{array} %
\usepackage{tabularx} %
\usepackage{xcolor} %
\usepackage{tabularx}
\usepackage{booktabs}

\usepackage{lipsum}  %
\usepackage{multicol} %
\usepackage{makecell} 
\usepackage{listings}
\usepackage{xcolor}

\newcommand{\fstar}{\ding{72}} 
\newcommand{\estar}{\ding{73}} 

\usepackage{tablefootnote}
\usepackage{hyperref}

\lstdefinelanguage{json}{
    basicstyle=\ttfamily\footnotesize,
    numbers=left,
    numberstyle=\tiny\color{gray},
    stepnumber=1,
    numbersep=8pt,
    showstringspaces=false,
    breaklines=true,
    frame=lines,
    backgroundcolor=\color{gray!10},
    morestring=[b]",
    literate=
     *{0}{{{\color{black}0}}}{1}
      {1}{{{\color{black}1}}}{1}
      {2}{{{\color{black}2}}}{1}
      {3}{{{\color{black}3}}}{1}
      {4}{{{\color{black}4}}}{1}
      {5}{{{\color{black}5}}}{1}
      {6}{{{\color{black}6}}}{1}
      {7}{{{\color{black}7}}}{1}
      {8}{{{\color{black}8}}}{1}
      {9}{{{\color{black}9}}}{1}
}

\makeatletter
\def\@BTrule[#1]{%
  \ifx\longtable\undefined
    \let\@BTswitch\@BTnormal
  \else\ifx\hline\LT@hline
    \nobreak
    \let\@BTswitch\@BLTrule
  \else
     \let\@BTswitch\@BTnormal
  \fi\fi
  \global\@thisrulewidth=#1\relax
  \ifnum\@thisruleclass=\tw@\vskip\@aboverulesep\else
  \ifnum\@lastruleclass=\z@\vskip\@aboverulesep\else
  \ifnum\@lastruleclass=\@ne\vskip\doublerulesep\fi\fi\fi
  \@BTswitch}
\makeatother

\addto\extrasenglish{
}

 {\begin{list}{}%
         {\setlength{\leftmargin}{#1}}%
         \item[]%
 }
 {\end{list}}

\reportnumber{001} %

\renewcommand{\today}{}

\title{\centering PaddleOCR-VL-1.5: Towards a Multi-Task 0.9B VLM for Robust In-the-Wild Document Parsing}

\author[*]{
\small
Cheng Cui, Ting Sun, Suyin Liang, Tingquan Gao, Zelun Zhang, 
\vspace{-0.4cm}
\\
\small
 Jiaxuan Liu, Xueqing Wang, Changda Zhou, Hongen Liu, Manhui Lin, 
\\
\small
Yue Zhang, Yubo Zhang,  Yi Liu, Dianhai Yu, Yanjun Ma
\vspace{0.2cm}
\\
\small
\textbf{PaddlePaddle Team, Baidu Inc.}
\\
\small
\texttt{paddleocr@baidu.com}
\vspace{0.2cm}
  \\
  {\small
  \raggedright{  
  \small
  \hspace{12.6em}  
  \includegraphics[height=1.0em]{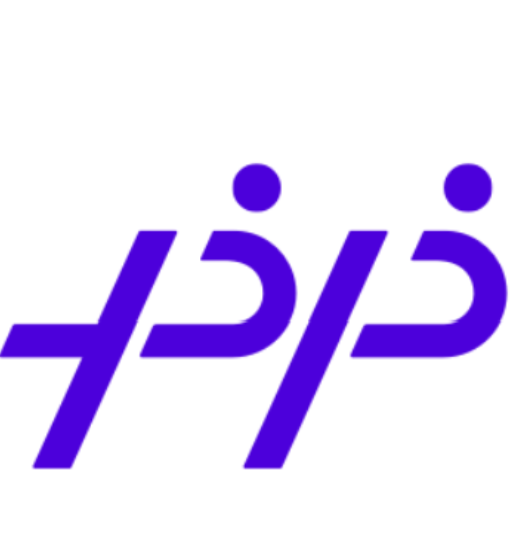} \textbf{Official Website}: \url{https://www.paddleocr.com} \\
  \hspace{-1.2em}  
  \small
  \hspace{6.55em}  
  \includegraphics[height=0.9em]{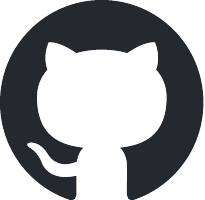} \textbf{Source Code}: \url{https://github.com/PaddlePaddle/PaddleOCR} \\
  \hspace{-1.2em}  
  \small
  \hspace{6.9em}  
  \includegraphics[height=1.0em]{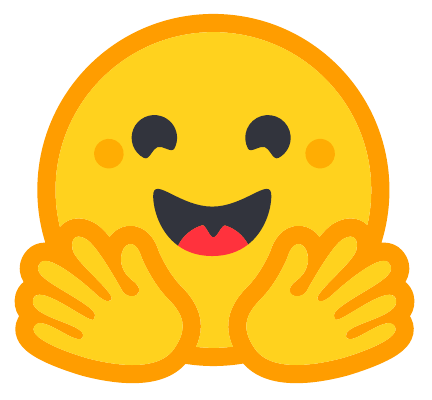} \textbf{Models}: \url{https://huggingface.co/PaddlePaddle} \\
  }

  }
}

\input{commands.tex}

\begin{abstract}

\vspace{-0.5cm} 

We introduce PaddleOCR-VL-1.5, an upgraded model achieving a new state-of-the-art (SOTA) accuracy of 94.5\% on OmniDocBench v1.5. To rigorously evaluate robustness against real-world physical distortions—including scanning, skew, warping, screen-photography
, and illumination—we propose the Real5-OmniDocBench benchmark. Experimental results demonstrate that this enhanced model attains SOTA performance on the newly curated benchmark. Furthermore, we extend the model's capabilities by incorporating seal recognition and text spotting tasks, while remaining a 0.9B ultra-compact VLM with high efficiency.

\end{abstract}

\begin{document}

\maketitle
\vspace{0cm} 
\begin{figure}[h]

\makebox[0pt][l]{\hspace{-0.5cm}\includegraphics[width=1.06\textwidth]{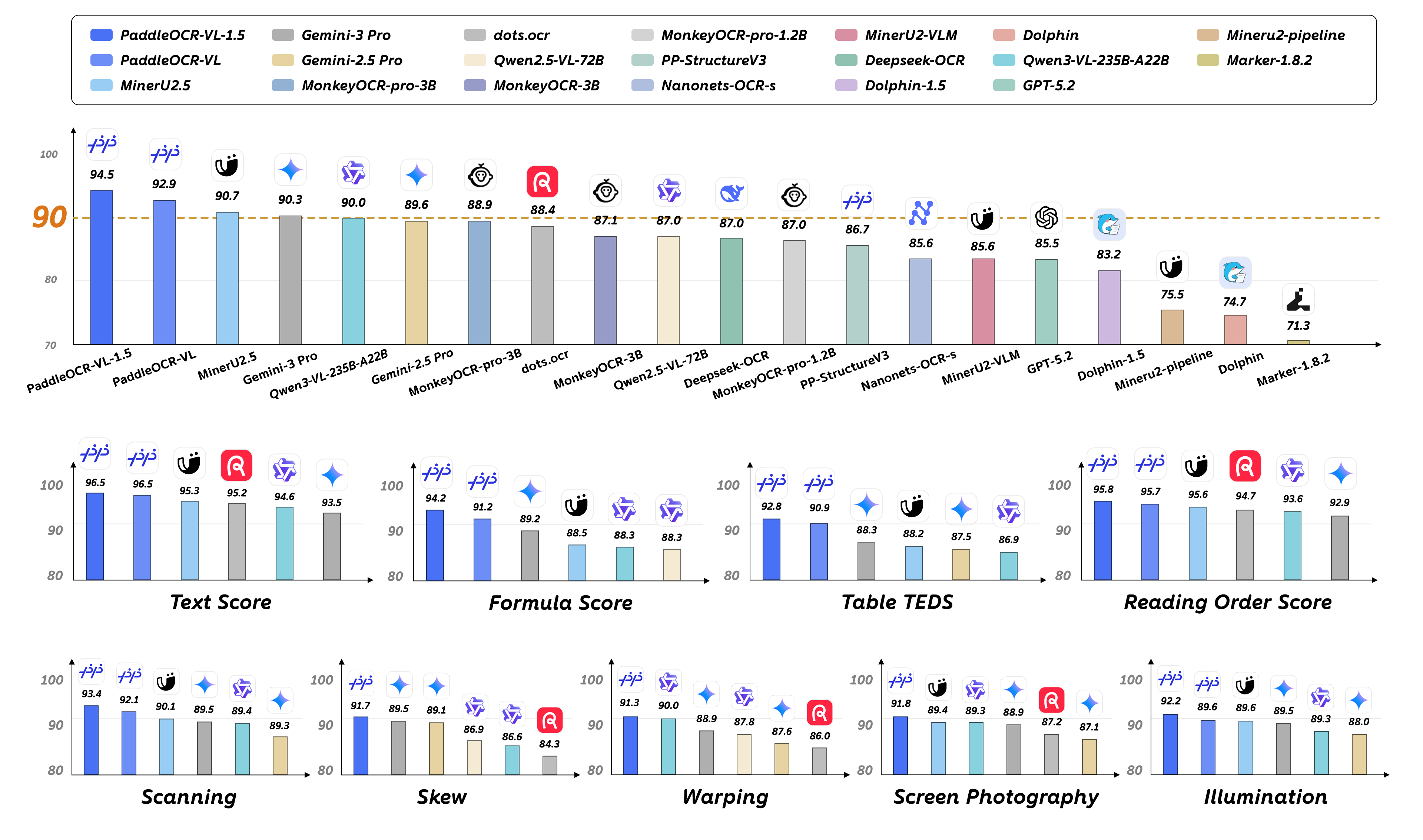}}

\caption{
    Performance of PaddleOCR-VL-1.5 on OmniDocBench v1.5 and Real5-OmniDocBench. 
}
\label{fig:dataset}
\end{figure}

\newpage
\setlength{\cftbeforesecskip}{6pt}   
\setlength{\cftbeforesubsecskip}{4pt} 
\setcounter{tocdepth}{2}
\tableofcontents

\newpage

\section{Introduction}

As the primary repository of human knowledge, documents are growing exponentially in both volume and complexity, establishing document parsing as a pivotal technology in the era of artificial intelligence. The ultimate objective of document parsing~\cite{li2025monkeyocr, niu2025mineru2, feng2025dolphin, liu2025points} extends beyond mere text recognition; it aims to reconstruct the deep structural and semantic layout of a document. By meticulously distinguishing text blocks, decoding complex formulas and tables, and deducing the logical reading order, advanced parsing lays the groundwork for Large Language Models (LLMs)~\cite{ernie2025technicalreport, yang2025qwen3, achiam2023gpt}. Crucially, this capability empowers Retrieval-Augmented Generation (RAG) systems~\cite{lewis2020retrieval} to ingest high-fidelity knowledge, thereby enhancing their reliability in downstream applications.

The field has witnessed a surge of innovation following October 2025, with several significant document parsing solutions emerging to push the boundaries of document intelligence. Notably, PaddleOCR-VL~\cite{cui2025paddleocrvl} established a high performance baseline, surpassing contemporary SOTA metrics with only 0.9 billion parameters and demonstrating strong multi-scenario generalization. Concurrently, DeepSeek-OCR~\cite{wei2025deepseek} leverages an optical 2D mapping methodology to enable high-ratio vision-to-text compression, offering robust end-to-end parsing capabilities. MonkeyOCR v1.5~\cite{zhang2025monkeyocrv15technicalreport} further enhances the three-stage parsing framework, while HunyuanOCR~\cite{hunyuanvisionteam2025hunyuanocrtechnicalreport} extends expert OCR capabilities through a unified architecture supporting translation and extraction.

Despite these advancements, a critical gap remains: most existing models are primarily optimized for "digital-born" or cleanly scanned documents. Real-world scenarios involving extreme physical distortions—such as aggressive skewing, non-rigid warping of pages, screen-capture moiré patterns, and erratic lighting—remain significant hurdles that even state-of-the-art solutions have yet to fully overcome.

To bridge this gap, we present PaddleOCR-VL-1.5, a high-performance, resource-efficient document parsing solution that significantly enhances both general precision and real-world robustness. Building upon the proven 0.9B ultra-compact architecture, PaddleOCR-VL-1.5 introduces several critical advancements:

\begin{itemize}
    \item Firstly, we upgrade the layout engine to PP-DocLayoutV3. Unlike previous Layout Analysis methods (e.g., Dolphin~\cite{feng2025dolphin}, MinerU2.5~\cite{niu2025mineru2}, or even PP-DocLayoutV2~\cite{sun2025pp}), PP-DocLayoutV3 is specifically engineered to handle non-planar document images. It can directly predict multi-point bounding boxes for layout elements—as opposed to standard two-point boxes—and determine logical reading orders for skewed and warped surfaces within a single forward pass, significantly reducing cascading errors.
    \item Secondly, we expand the model's core capabilities. While maintaining the efficient NaViT-style dynamic resolution encoder and the ERNIE-4.5-0.3B~\cite{ernie2025technicalreport} language backbone, we have integrated new tasks including seal recognition and text spotting. Systematic optimizations in text, table, and formula recognition have further propelled the model to a new performance milestone.
    \item Thirdly, we construct Real5-OmniDocBench to evaluate in-the-wild robustness. Recognizing the lack of benchmarks for physical distortions, we curated this dataset based on OmniDocBench v1.5~\cite{ouyang2025omnidocbench}. It comprises five distinct scenarios: scanning, warping, screen photography, illumination, and skew. By maintaining a strict one-to-one correspondence with the original ground-truth annotations, Real5-OmniDocBench serves as a rigorous benchmark for assessing model resilience in practical applications.
\end{itemize}

Comprehensive benchmarking confirms that PaddleOCR-VL-1.5 establishes a new state-of-the-art (SOTA) standard. On the OmniDocBench v1.5 benchmark, our model achieves a breakthrough accuracy of 94.5\%, maintaining its position as the official top-ranked solution. More importantly, on the newly curated Real5-OmniDocBench, the model sets a new record with an overall accuracy of 92.05\%. Despite its compact 0.9B scale, it significantly outperforms massive general VLMs, such as Qwen3-VL-235B~\cite{yang2025qwen3} and Gemini-3 Pro~\cite{gemini30}, highlighting its exceptional parameter efficiency. Furthermore, our model expands its capabilities to text spotting and seal recognition, attaining leading performance across diverse and challenging benchmarks. These results collectively validate its superior robustness and generalization in complex, real-world scenarios. Appendix \ref{sec:Comparison of PaddleOCR-VL-1.5 and 1.0 Models} details the specific upgrades and changes in PaddleOCR-VL-1.5 compared to its predecessor.

\section{PaddleOCR-VL-1.5}

\subsection{Architecture}
\label{Architecture PaddleOCR-VL}

\begin{figure}[H]
\centering
\includegraphics[width=\linewidth]{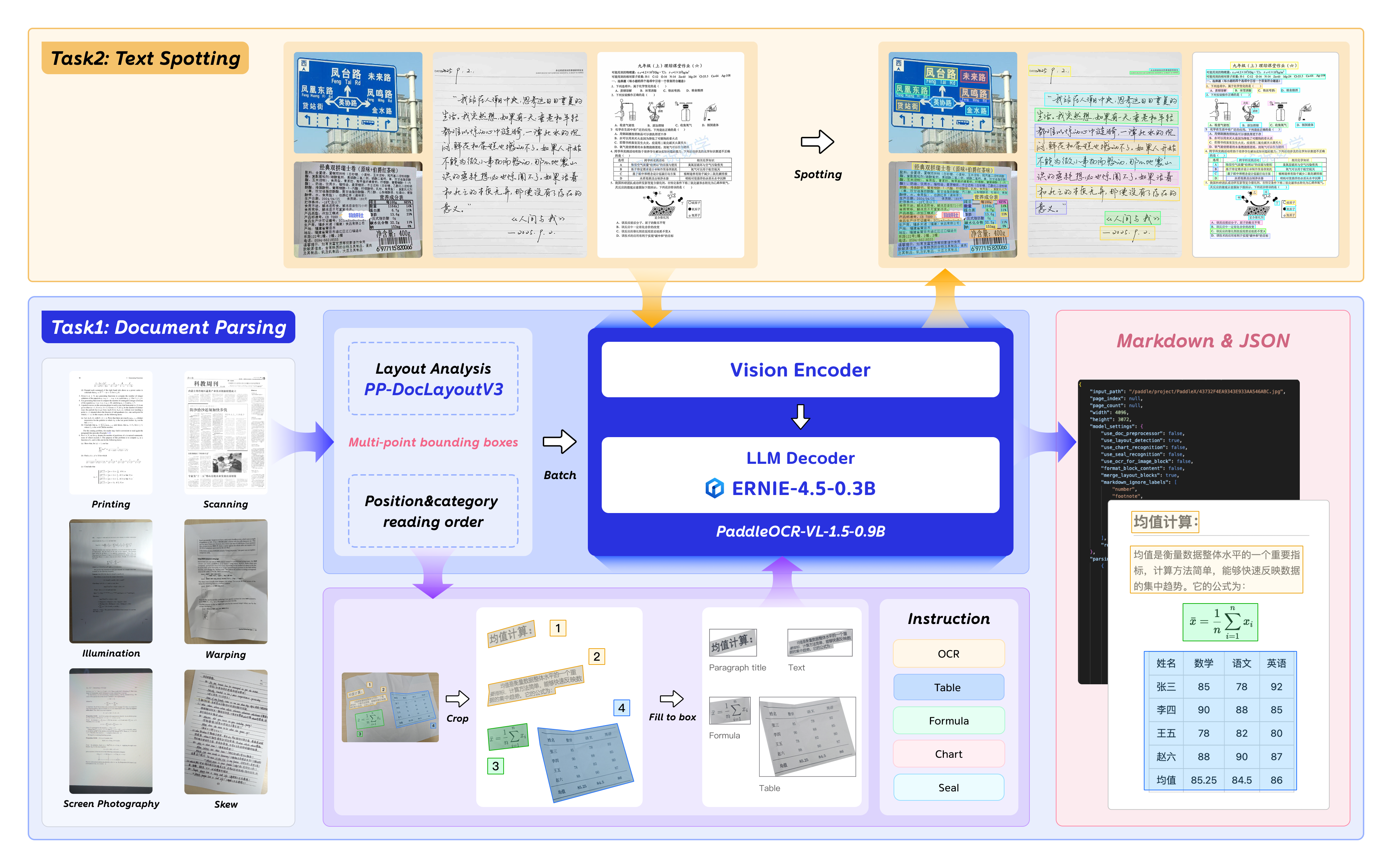} 

\caption{
    \centering
    The overview of PaddleOCR-VL-1.5.
}
\label{fig:model_overview}
\end{figure}

PaddleOCR-VL-1.5 introduces an enhanced framework capable of handling both Document Parsing and Text Spotting, as depicted in Figure~\ref{fig:model_overview}.

For the Document Parsing task, PaddleOCR-VL-1.5 adopts a robust two-stage framework. In the initial stage, PP-DocLayoutV3 performs sophisticated layout analysis. Beyond standard axis-aligned detection, it is specifically optimized for real-world complexity by employing multi-point localization (e.g., quadrilaterals or polygons). This allows for the precise boundary anchoring of semantic regions even under severe perspective tilt or physical curvature, while simultaneously establishing the logical reading order. In the second stage, the PaddleOCR-VL-1.5-0.9B model takes these geometrically-rectified or localized regions as input to perform high-fidelity recognition across diverse modalities, including text, complex tables, mathematical formulas, charts and seals. To conclude the pipeline, a lightweight post-processing engine orchestrates these outputs into structured formats such as Markdown and JSON, while providing advanced capabilities such as cross-page table merging and heading hierarchy refinement.

For the Spotting task, the framework simplifies its workflow by directly utilizing the PaddleOCR-VL-1.5-0.9B model for end-to-end text detection and recognition. This approach enables end-to-end text detection and recognition across a wide spectrum of domains—ranging from standard documents, identification cards, and ancient manuscripts to unconstrained scenarios like advertising posters, dialogue screenshots, signboards, and multilingual texts.

\subsubsection{PP-DocLayoutV3: Unified Layout Analysis}

To address the challenges of complex physical distortions—including skew, warping, and illumination—and to overcome the high latency inherent in autoregressive Vision-Language Models (VLMs), we introduce PP-DocLayoutV3. This version represents a significant architectural evolution from its predecessor by transitioning from standard rectangular detection to a robust instance segmentation framework, while simultaneously integrating reading order prediction into a unified, end-to-end Transformer architecture.

Building upon the high-efficiency RT-DETR object detector~\cite{zhao2024detrs}, PP-DocLayoutV3 adopts a mask-based detection head. This allows the model to predict precise, pixel-accurate masks for layout elements rather than simple bounding boxes. Such a capability is critical for isolating document components in non-ideal scenarios, such as skewed or warped pages, where traditional axis-aligned boxes frequently overlap or capture excessive background noise.

Unlike the decoupled pointer network employed in PP-DocLayoutV2~\cite{cui2025paddleocrvl}, PP-DocLayoutV3 integrates Reading Order Prediction directly into the Transformer decoder layers. By merging detection, segmentation, and ordering into a single vision-centric model, PP-DocLayoutV3 eliminates the need for redundant post-processing and separate feature extraction steps. 

\begin{figure}[ht]
\centering
\includegraphics[width=\linewidth]{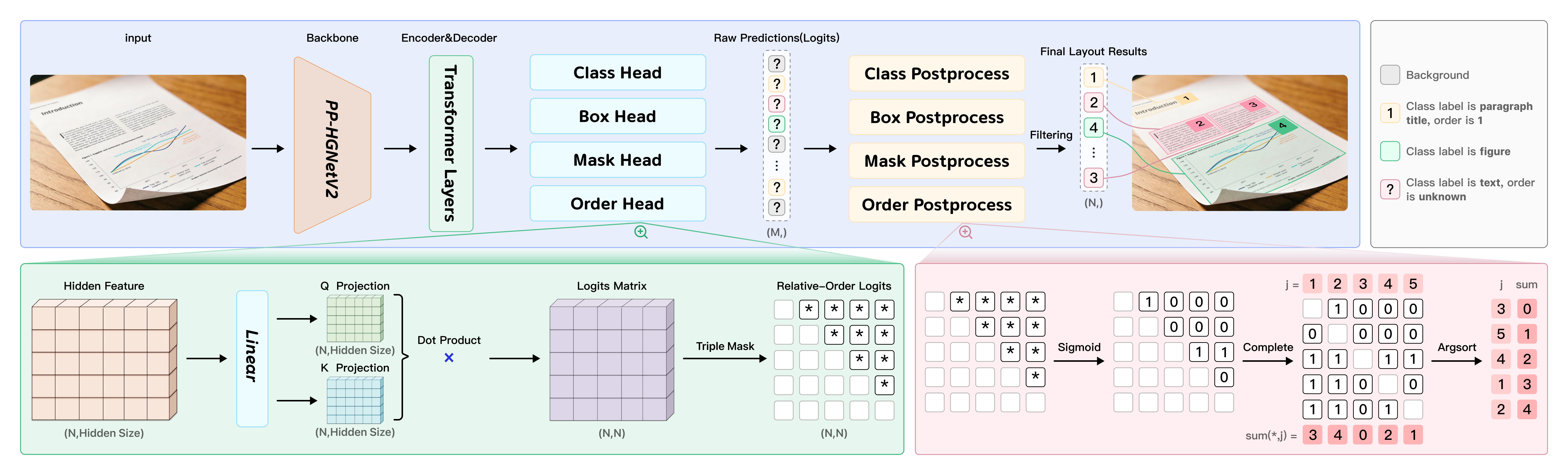}
\caption{The unified architecture of PP-DocLayoutV3, featuring parallel heads for instance segmentation and relational reading order prediction.}
\label{fig:PP-DocLayoutV3}
\end{figure}

The core architectural innovation of PP-DocLayoutV3 is the integration of Reading Order Prediction directly into the Transformer decoder. Specifically, our model extends the RT-DETR framework to simultaneously optimize geometric localization and logical sequencing. Following the query-based paradigm, the decoder iteratively refines $N$ object queries $Q = \{q_i\}_{i=1}^N \in \mathbb{R}^{N \times d}$. The reading order is then derived from the refined query embeddings of the final decoder layer through a Global Pointer Mechanism.

We project the refined queries into a shared relational space to compute the pairwise precedence score $S_{i,j}$:

\begin{equation}
    S_{i,j} = \frac{f(q_i, q_j) - f(q_j, q_i)}{\sqrt{d_h}}, \quad \text{where } f(q_i, q_j) = (W_q q_i)^\top (W_k q_j)
\end{equation}

where $W_q, W_k \in \mathbb{R}^{d \times d_h}$ are learnable projection matrices and $d_h$ denotes the hidden dimension. The resulting relation matrix $S \in \mathbb{R}^{N \times N}$ is constrained to be anti-symmetric such that $S_{i,j} = -S_{j,i}$, where $S_{i,j} > 0$ implies element $i$ precedes element $j$.

During inference, to derive a globally consistent sequence from these pairwise relations, we implement a Voting-based Ranking strategy. We first apply the sigmoid function $\sigma(\cdot)$ to the relation matrix $S$ and mask the diagonal elements. The absolute precedence votes $V_j$ for each element $j$ is computed by aggregating the probabilities of other elements preceding it:

\begin{equation}
    V_j = \sum_{i=1, i \neq j}^{N} \sigma(S_{i,j}).
\end{equation}

The final reading order is determined by sorting the elements in ascending order of their total votes $V_j$. This joint optimization ensures that the logical sequence is highly sensitive to the refined object features, leading to superior performance on complex, multi-column, and non-standard document layouts.

By merging detection, segmentation, and ordering into a single vision-centric model, PP-DocLayoutV3 eliminates the need for redundant post-processing and separate feature extraction. The model produces the complete document structure in a single forward pass, where the multi-head system concurrently outputs classification labels, bounding box coordinates, pixel-accurate segments, and the logical reading sequence.

\subsubsection{PaddleOCR-VL-1.5-0.9B: Element-level Recognition and Text Spotting}
\label{PaddleOCR-VL-1.5-0.9B arch}

The PaddleOCR-VL-1.5-0.9B inherits the lightweight architecture of PaddleOCR-VL-0.9B~\cite{cui2025paddleocrvl}, integrating a Native Resolution Visual Encoder~\cite{dehghani2023patch}, an Adaptive MLP Connector, and the Lightweight ERNIE-4.5-0.3B Language Model~\cite{ernie2025technicalreport}. In this update, the model's capabilities have been expanded to include Seal Recognition and Text Spotting. Consequently, the model now supports a comprehensive set of six core tasks: OCR, Formula Recognition, Table Recognition, Chart Recognition, Seal Recognition, and Text Spotting. 

Compared to its predecessor, PaddleOCR-VL-1.5-0.9B demonstrates significant enhancements in recognition accuracy for complex tables and mathematical formulas. Furthermore, the model incorporates finer-grained optimizations for rare characters, ancient Chinese texts, multilingual tables, and text decorations such as underlines and emphasis marks.

\subsection{Training Recipe}

 The following sections introduce the training details of these two modules: PP-DocLayoutV3 for layout analysis and PaddleOCR-VL-1.5-0.9B for element recognition and text spotting.

\subsubsection{Layout Analysis}

\label{Layout Analysis}

The training of PP-DocLayoutV3 evolves from the two-stage decoupled process used in PP-DocLayoutV2~\cite{cui2025paddleocrvl} to a more sophisticated end-to-end joint optimization strategy. This approach allows the detection, instance segmentation, and reading order modules to share a unified feature representation, leading to better alignment between spatial localization and logical sequencing.

The model is initialized with the pre-trained weights of PP-DocLayout\_plus-L~\cite{sun2025pp}, we scaled our training corpus to over 38k high-quality document samples. Each sample underwent rigorous manual annotation to provide ground truth, include the coordinates, categorical label and absolute reading order for every layout elements.

To achieve the environmental robustness, we designed a specialized Distortion-Aware Data Augmentation pipeline. Unlike standard augmentations, this pipeline specifically simulates complex physical deformations found in real-world mobile photography.

We utilize the AdamW optimizer with a weight decay of $0.0001$. The learning rate is set to a constant $2 \times 10^{-4}$ to ensure stable convergence of the integrated Global Pointer and Mask heads. The model is trained for 150 epochs with a total batch size of 32. 

In contrast to the previous version, all components—including the RT-DETR backbone and the integrated reading order transformer—are trained simultaneously. This end-to-end supervision ensures that the learned queries in the Transformer decoder capture both the geometric boundaries and the topological relationships of the document elements.

\subsubsection{Element-level Recognition and Text Spotting}
\label{training_strategy}

Building upon the architecture described in Section \ref{PaddleOCR-VL-1.5-0.9B arch}, PaddleOCR-VL-1.5-0.9B introduces a progressive training paradigm using PaddleFormers~\cite{PaddlePaddle_PaddleFormers}, which is a high-performance training toolkit for LLMs and VLMs built on the PaddlePaddle framework~\cite{ma2019paddlepaddle}. While we retain the effective post-adaptation strategy and initialization settings from our previous version, the training methodology has been significantly upgraded to enhance data scale, task diversity, and model robustness. The overview of the three stages is presented in Table \ref{training}.

\begin{table}[h]
\centering
 \fontsize{8}{8}\selectfont
\renewcommand{\arraystretch}{1.2}
\begin{tabular}{l|cc}
\toprule
\textbf{Settings} & \textbf{Pre-training} & \textbf{Post-training}  \\  \midrule
Training Samples & 46M  & 5.6M  \\
Max Resolution & 1280 $\times$ 28 $\times$ 28 & 1280 $\times$ 28 $\times$ 28  \\
Sequence length & 16384 & 16384 \\
Trainable components & All & All \\
Batch sizes & 128 & 128  \\
Data Augmentation & Yes & Yes  \\
Maximum LR & $5 \times 10^{-5}$ & $8 \times 10^{-6}$  \\
Epoch & 1 & 1  \\ \bottomrule
\end{tabular}
\caption{Training settings for PaddleOCR-VL-1.5-0.9B.}
\label{training}
\end{table}

\textbf{Pre-training: Enhanced Vision-Language Alignment.} 
While the fundamental objective remains aligning visual features with textual semantics, this stage undergoes a substantial data upgrade compared to PaddleOCR-VL-0.9B~\cite{cui2025paddleocrvl}, scaling the pre-training dataset from 29 million to 46 million image-text pairs. This expansion represents a qualitative leap in data distribution rather than a mere quantitative increase. Specifically, to enhance the generalization of the visual backbone and support the newly introduced capabilities, we incorporate a broader spectrum of multilingual documents and complex real-world scenarios. Furthermore, we intentionally inject large-scale pre-training data related to seal recognition and text spotting during this alignment phase. Specifically, the maximum resolution for the spotting task is increased to $2048 \times 28 \times 28$ pixels, enabling the model to achieve more precise localization and recognition of text. By introducing these task-specific priors early in the training pipeline, the model establishes a robust foundation capable of capturing intricate visual patterns and effectively supporting the fine-grained localization and recognition tasks required in subsequent stages.

\textbf{Post-training: Instruction Fine-tuning with New Capabilities.} 
In this stage, we inherit the four fundamental instruction tasks from PaddleOCR-VL-0.9B—\textbf{OCR, Table, Formula, and Chart Recognition}—ensuring backward compatibility and high performance on standard document elements. The key innovation in PaddleOCR-VL-1.5-0.9B lies in the addition of two specialized tasks:

\begin{enumerate}[label=\arabic*.]
    \item \textbf{Seal Recognition:} We introduce a specific instruction to handle official seals and stamps, addressing challenges such as curved text, blur images and background interference.
        
    \item \textbf{Text Spotting (Grounded OCR):} 
    Unlike standard OCR, which solely outputs textual content, the text spotting task requires the model to simultaneously predict the text and its precise spatial location following the natural reading order. To accommodate complex layouts found in real-world scenarios (e.g., rotated text, common scene, or dense forms), we adopt a \textbf{4-point quadrilateral representation} rather than the traditional 2-point bounding box. The 4-point format defines a text region using four vertices: Top-Left (TL), Top-Right (TR), Bottom-Right (BR), and Bottom-Left (BL). This formulation provides superior flexibility in localizing inclined and irregular text shapes that a standard axis-aligned rectangle cannot tightly enclose.
    
    Formally, for a given text instance, the target sequence is constructed by appending eight location tokens to the text tokens:
    \begin{equation}
        Y = \text{Text} \oplus \texttt{<LOC\_}x_{\text{TL}}\texttt{>} \texttt{<LOC\_}y_{\text{TL}}\texttt{>} \dots \texttt{<LOC\_}y_{\text{BL}}\texttt{>}
    \end{equation}
    Here, we introduce a set of tokens $\{\texttt{<LOC\_0>}, \dots, \texttt{<LOC\_1000>}\}$ to the model's vocabulary to represent normalized coordinates. Unlike treating coordinates as plain numerical text, these dedicated special tokens allow the model to learn specific embeddings for spatial information and prevent tokenization fragmentation.
    
    For example, a recognized instance of the word "DREAM" is represented as:
    \begin{quote}
    \small
        \texttt{DREAM} \texttt{<LOC\_253>} \texttt{<LOC\_286>} \texttt{<LOC\_346>} \texttt{<LOC\_298>} \texttt{<LOC\_345>} \texttt{<LOC\_339>} \texttt{<LOC\_252>} \texttt{<LOC\_330>}
    \end{quote}
    This unified representation enables the model to perform end-to-end recognition and fine-grained localization within a single generation pass.
\end{enumerate}

To enhance generalization and unify diverse label styles, we introduce a Reinforcement Learning stage leveraging Group Relative Policy Optimization(GRPO)~\cite{shao2024deepseekmath}. By executing parallel rollouts and calculating relative advantages within each group, GRPO facilitates robust policy updates and mitigates style inconsistency. This process is supported by a dynamic data screening protocol that prioritizes challenging samples with high reward potential and entropy uncertainty, ensuring the model focuses on non-trivial, high-value learning cases.

\section{Dataset}

\subsection{Layout Analysis}

To ensure robust model performance across diverse real-world document scenarios, we curate a in-house dataset for layout analysis. The data sources encompass 38k document images across diverse domains, including Academic papers, Textbooks, Market Analysis, Financial reports, Slides, Newspapers, Supplementary Teaching Materials, Examination Papers, and various Invoices and Receipts. The dataset features meticulous manual annotations across 25 distinct component categories: Paragraph Title, Image, Text, Number, Abstract, Content, Figure Title, Display Formula, Table, Reference, Doc Title, Footnote, Header, Algorithm, Footer, Seal, Chart, Formula Number, Aside Text, Reference Content, Header Image, Footer Image, Inline Formula, Vertical Text, and Vision Footnote. All documents are manually annotated with element-level boundaries and their corresponding reading order, enabling effective training and evaluation for both layout element detection and reading order restoration. This high-quality ground truth ensures that the model can accurately reconstruct both the spatial structure and the logical flow of complex documents.

The data curation process incorporates specific data mining strategies aimed at expanding dataset diversity and identifying hard cases to improve model robustness. This workflow begins with clustering-based sampling applied to an extensive internal data pool, utilizing visual features to ensure a representative distribution and minimize redundancy. Subsequently, a hard-case mining pipeline is executed using PP-DocLayoutV2~\cite{cui2025paddleocrvl} for dual-threshold inference. Samples exhibiting a significant discrepancy in detection density between high and low confidence thresholds are categorized as unstable cases. This methodology facilitates the systematic discovery of non-conventional layout structures—including comics, CAD drawings, and high-aspect ratio screenshots—which diverge from standard document formats. These instances are further refined through a human-in-the-loop process. Integrating these diverse scenarios into the dataset broadens the model's representation of characteristics and enhances its adaptive capacity in complex real-world document domains.

\subsection{PaddleOCR-VL-1.5-0.9B}

The data construction strategy for PaddleOCR-VL-1.5-0.9B is driven by two core objectives: enhancing model robustness on challenging samples and expanding the breadth of supported capabilities. Consequently, our data preparation pipeline is divided into two distinct parts: 
(1) \textbf{Hard Example Mining} (Section \ref{hard_example_mining}), which focuses on identifying and weighting high-uncertainty samples to refine the model's decision boundaries; and 
(2) \textbf{New Capability Data Construction} (Section \ref{new_capability_data}), which involves curating specialized datasets to unlock new skills such as text spotting, seal recognition, and advanced multilingual support.

\subsubsection{Data Selection Strategy: Uncertainty-Aware Cluster Sampling}
\label{hard_example_mining}

To maximize the efficiency of the instruction fine-tuning stage (Stage 2), we propose a data curation strategy designed to balance visual diversity and sample difficulty. Instead of uniform random sampling, we employ an \textbf{Uncertainty-Aware Cluster Sampling (UACS)} mechanism. This approach ensures that the training data covers a wide spectrum of visual scenarios while allocating more training budget to "hard" cases where the model exhibits high uncertainty.

\textbf{1. Visual Feature Clustering.} 
First, to guarantee the diversity of visual layouts across the six tasks (OCR, Table, Formula, Chart, Seal, and Spotting), we utilize the CLIP~\cite{radford2021learning} visual encoder to extract high-dimensional semantic embeddings for all candidate images. For each task, we apply K-Means clustering to partition the dataset $\mathcal{D}$ into $K$ distinct visual clusters $\{C_1, C_2, \dots, C_K\}$. This step groups samples with similar visual structures (e.g., solid line tables vs. wireless tables) together.

\textbf{2. Uncertainty Estimation.}
For each cluster $C_i$, we estimate its difficulty by measuring the model's prediction uncertainty. Specifically, we randomly sample a subset of images from $C_i$ and perform multiple inference passes with stochastic decoding using the pre-trained model from Stage 1. We calculate an uncertainty score $S_i$ based on the divergence of the generated outputs. A higher $S_i$ indicates that the model is inconsistent or unconfident regarding the samples in this cluster.

\textbf{3. Weighted Sampling Plan.} 
Based on the uncertainty score, we formulate a sampling plan to determine the number of samples $N_i$ to draw from each cluster $C_i$. Inspired by the principle of hard example mining, we adopt a polynomial weighting scheme to amplify the focus on harder clusters. Specifically, the allocated sample count $N_i$ for cluster $C_i$ is determined by:

\begin{equation}
    N_i = \min \left( \left\lfloor \frac{(S_i + \alpha)^\beta}{\sum_{j=1}^{K} (S_j + \alpha)^\beta} \times N_{\text{total}} \right\rfloor, |C_i| \right)
\end{equation}

where $S_i$ is the average uncertainty score of cluster $C_i$, and $|C_i|$ denotes the total number of available samples in that cluster. The parameters $\alpha$ and $\beta$ are the smoothing and power factors, respectively (set to $\alpha=1.0, \beta=2.0$ based on empirical observations). $N_{\text{total}}$ represents the total sampling budget. This strategy allows us to dynamically up-sample complex scenarios (e.g., distorted seals, dense tables) while maintaining a representative baseline for simpler cases.

As detailed in the previous section, we employ the Uncertainty-Aware Cluster Sampling (UACS) strategy to select the most effective training samples based on visual clustering and inference variance.

\subsubsection{Data Construction for New Capabilities}
\label{new_capability_data}

In addition to data quality control, we expanded the VLM's capabilities by integrating data spanning a wider array of tasks, languages, and document types. This expansion focuses on these key dimensions: \textbf{Spotting}, \textbf{Specialized Text (Seals)}, \textbf{OCR Enhancement}, and \textbf{Complex Tables, Formulas, and Chart}.

\textbf{Spotting:} We collected a large-scale and diverse set of images covering a wide range of real-world scenarios, such as financial research reports, tabular documents, handwritten materials, classical texts, and other complex document and natural scene images.  During the annotation process, we jointly employed PP-OCRv5~\cite{cui2025paddleocr} to generate initial recognition results, followed by an IoU-based cross-filtering strategy to eliminate low-quality and inconsistent samples, while for a subset of samples with ambiguous or inaccurate labels, multimodal models including PaddleOCR-VL~\cite{cui2025paddleocrvl} and Qwen3-VL~\cite{yang2025qwen3} were further leveraged for label refinement, thereby substantially improving the overall annotation quality and robustness of the dataset.

\textbf{Seal:} We combined synthetic and real-world images of contracts, invoices, and commemorative seals to build a high-quality dataset. Labels were generated using Qwen3-VL~\cite{yang2025qwen3} and refined through a fine-tuning-based re-labeling process. Challenging cases were manually corrected to ensure final annotation accuracy and robustness.

\textbf{OCR:} We have significantly enhanced the model’s capabilities by refining the dataset’s precision and enlarging the functional scope. This includes systematically correcting formula representations and line-breaking logic, alongside extending support for education-specific markings like emphasis dots and underlines to capture instructional semantics. Furthermore, the integration of Bengali and China’s Tibetan scripts broadens the model’s linguistic versatility, ensuring robust performance across diverse writing systems and educational contexts.

\textbf{Formula:} Formula dataset incorporates CV-simulated artifacts, such as Gaussian illumination and harmonic moiré, to replicate physical conditions like scanning, warping, screen-capture, and geometric skewing. These samples encompass a wide spectrum of environmental variables, including light fluctuations and complex document distortions encountered in realistic scenarios.

\textbf{Table:} Tables has been expanded to cover an extensive range of scenarios, including financial reports, academic papers, and complex industrial forms. Integration of diverse structures, such as registration and catalog tables. A key focus is placed on the precise recognition of cell-level formulas and multilingual content within dense table environments. These advancements ensure high-fidelity conversion into structured formats, even when handling intricate cell structures and professional notations.

\section{Evaluation}

\label{sec:experiments} 

To thoroughly assess the effectiveness of PaddleOCR-VL-1.5, we conducted evaluations on the document parsing benchmark OmniDocBench v1.5 and its derived real-world dataset, Real5-OmniDocBench. Furthermore, we expanded the evaluation scope by incorporating text spotting and seal recognition tasks to provide a more comprehensive analysis of the model's performance in practical and complex scenarios.

\subsection{Document Parsing}
\label{Document Parsing}

This section details the evaluation of end-to-end document parsing capabilities using the following two benchmarks, aiming to measure its overall performance in real-world document scenarios. 

\paragraph{OmniDocBench v1.5} To comprehensively evaluate the document parsing capabilities, we conducted extensive experiments on the OmniDocBench v1.5~\cite{niu2025mineru2} benchmark. It is an expansion of version v1.0, adding 374 new documents for a total of 1,355 document pages. It features a more balanced distribution of data in both Chinese and English, as well as a richer inclusion of formulas and other elements. Compared to version v1.0, the evaluation method has been updated. While text and reading order are still evaluated using Edit Distance, and tables are evaluated using Tree-Edit-Distance-based Similarity (TEDS), formulas are now assessed using the Character Detection Matching (CDM) \cite{Wang_cdm_2025_CVPR} method. This metric provides a more objective and robust evaluation of the correctness of predicted formulas. The overall metric is a weighted combination of the metrics for text, formulas, and tables. 

Table~\ref{tab:omni15_performance} demonstrates that PaddleOCR-VL-1.5 achieves SOTA performance, consistently outperforming existing pipeline tools, general VLMs, and specialized document parsing models across all key metrics. Notably, PaddleOCR-VL-1.5 exhibits a substantial performance leap over its predecessor, PaddleOCR-VL, raising the overall score from 92.86\% to a top-ranking 94.50\%. Specifically, it achieves increases of 2.99\%, 1.87\%, and 0.1\% in the CDM Score, Table-TEDS, and Reading Order scores, respectively. Furthermore, our model establishes new SOTA results in all sub-tasks, including a reduced Text-Edit distance of 0.035, an improved Formula-CDM score of 94.21\%, and leading scores of 92.76\% and 95.79\% in Table-TEDS and Table-TEDS-S, respectively. These improvements, particularly in maintaining a high reading ordering score of 0.042, underscore the model's enhanced precision in text recognition, formula extraction, and complex table structure analysis.

\begin{table}[!t]
    \centering

    \resizebox{\textwidth}{!}{%
    \renewcommand{\arraystretch}{1.2}
    \begin{tabular}{l|ll|c|c c c c c}
        \toprule
        \textbf{Model Type} & \textbf{Methods} & \textbf{Parameters} & \textbf{Overall$\uparrow$} & \textbf{Text\textsuperscript{Edit}$\downarrow$} & \textbf{Formula\textsuperscript{CDM}$\uparrow$} & \textbf{Table\textsuperscript{TEDS}$\uparrow$} & \textbf{Table\textsuperscript{TEDS-S}$\uparrow$} & \textbf{Reading Order\textsuperscript{Edit}$\downarrow$} \\    \midrule
        \multirow{3}{*}{\textbf{Pipeline Tools}} & Marker-1.8.2~\cite{vik2024marker} & - & 71.30 & 0.206 & 76.66 & 57.88 & 71.17 & 0.250 \\
        & Mineru2-pipeline~\cite{MinerU2} & - & 75.51 & 0.209 & 76.55 & 70.90 & 79.11 & 0.225 \\
        & PP-StructureV3~\cite{cui2025paddleocr} & - & 86.73 & 0.073 & 85.79 & 81.68 & 89.48 & 0.073 \\
        \midrule
        \multirow{8}{*}{\textbf{General VLMs}} & GPT-4o~\cite{achiam2023gpt} & - & 75.02 & 0.217 & 79.70 & 67.07 & 76.09 & 0.148 \\
        & InternVL3-76B~\cite{zhu2025internvl3} & 76B & 80.33 & 0.131 & 83.42 & 70.64 & 77.74 & 0.113 \\
        & InternVL3.5-241B~\cite{wang2025internvl35} & 241B & 82.67 & 0.142 & 87.23 & 75.00 & 81.28 & 0.125 \\
        & GPT-5.2~\cite{gpt5_2} & - & 85.50 & 0.123 & 86.11 & 82.66 & 87.35 & 0.099 \\
        & Qwen2.5-VL-72B~\cite{bai2025qwen2} & 72B & 87.02 & 0.094 & 88.27 & 82.15 & 86.22 & 0.102 \\
        & Gemini-2.5 Pro~\cite{gemini25} & - & 88.03 & 0.075 & 85.82 & 85.71 & 90.29 & 0.097 \\
        & Qwen3-VL-235B-A22B-Instruct~\cite{yang2025qwen3} & 235B & 89.15 & 0.069 & 88.14 & 86.21 & 	90.55 & 0.068 \\
        & Gemini-3 Pro~\cite{gemini30} & - & 90.33 & 0.065 & 89.18 & 88.28 & 90.29 & 0.071 \\
        \midrule
        \multirow{16}{*}{\textbf{Specialized VLMs}} & Dolphin~\cite{feng2025dolphin} & 0.3B & 74.67 & 0.125 & 67.85 & 68.70 & 77.77 & 0.124 \\
        & OCRFlux-3B~\cite{OCRFlux2025} & 3B & 74.82 & 0.193 & 68.03 & 75.75 & 80.23 & 0.202 \\
        & Mistral OCR~\cite{mistral} & - & 78.83 & 0.164 & 82.84 & 70.03 & 78.04 & 0.144 \\
        & POINTS-Reader~\cite{liu2025points} & 3B & 80.98 & 0.134 & 79.20 & 77.13 & 81.66 & 0.145 \\
        & olmOCR-7B~\cite{poznanski2025olmocr} & 7B & 81.79 & 0.096 & 86.04 & 68.92 & 74.77 & 0.121 \\
        & Dolphin-1.5 ~\cite{feng2025dolphin} & 0.3B & 83.21 & 0.092 & 80.78 & 78.06 & 84.10 & 0.080 \\
        & MinerU2-VLM~\cite{MinerU2} & 0.9B & 85.56 & 0.078 & 80.95 & 83.54 & 87.66 & 0.086 \\
        & Nanonets-OCR-s~\cite{Nanonets-OCR-S} & 3B & 85.59 & 0.093 & 85.90 & 80.14 & 85.57 & 0.108 \\
           & MonkeyOCR-pro-1.2B~\cite{li2025monkeyocr} & 1.9B & 86.96 & 0.084 & 85.02 & 84.24 & 89.02 & 0.130 \\
        & Deepseek-OCR~\cite{wei2025deepseek} & 3B & 87.01 & 0.073 & 83.37 & 84.97 & 88.80 & 0.086 \\
        & MonkeyOCR-3B~\cite{li2025monkeyocr} & 3.7B & 87.13 & 0.075 & 87.45 & 81.39 & 85.92 & 0.129 \\
        & dots.ocr~\cite{dotsocr} & 3B & 88.41 & 0.048 & 83.22 & 86.78 & 90.62 & 0.053 \\
      & MonkeyOCR-pro-3B~\cite{li2025monkeyocr} & 3.7B & 88.85 & 0.075 & 87.25 & 86.78 & 90.63 & 0.128 \\
        & MinerU2.5~\cite{niu2025mineru2} & 1.2B & 90.67 & \cellcolor{cyan!15}\underline{0.047} & 88.46 & 88.22 & 92.38 & 0.044 \\ 

        & \textbf{PaddleOCR-VL~\cite{cui2025paddleocrvl}} & 0.9B &\cellcolor{cyan!15}\underline{92.86} & \cellcolor{red!15}\textbf{0.035} & \cellcolor{cyan!15}\underline{91.22} & \cellcolor{cyan!15}\underline{90.89} &\cellcolor{cyan!15}\underline{94.76} & \cellcolor{cyan!15}\underline{0.043} \\
        
        & \textbf{PaddleOCR-VL-1.5} & 0.9B & \cellcolor{red!15}\textbf{94.50} & \cellcolor{red!15}\textbf{0.035} & \cellcolor{red!15}\textbf{94.21} & \cellcolor{red!15}\textbf{92.76} & \cellcolor{red!15}\textbf{95.79} & \cellcolor{red!15}\textbf{0.042} \\
        \bottomrule
    \end{tabular}%
    }
   \caption{Comprehensive evaluation on OmniDocBench v1.5. Performance metrics are cited from the official leaderboard \cite{ouyang2025omnidocbench}, except for Gemini-3 Pro, GPT-5.2, Qwen3-VL-235B-A22B-Instruct and our model, which were evaluated independently.}
       \label{tab:omni15_performance}
\end{table}

\paragraph{Real5-OmniDocBench}

\begin{figure}[h]
\centering
\includegraphics[width=\linewidth]{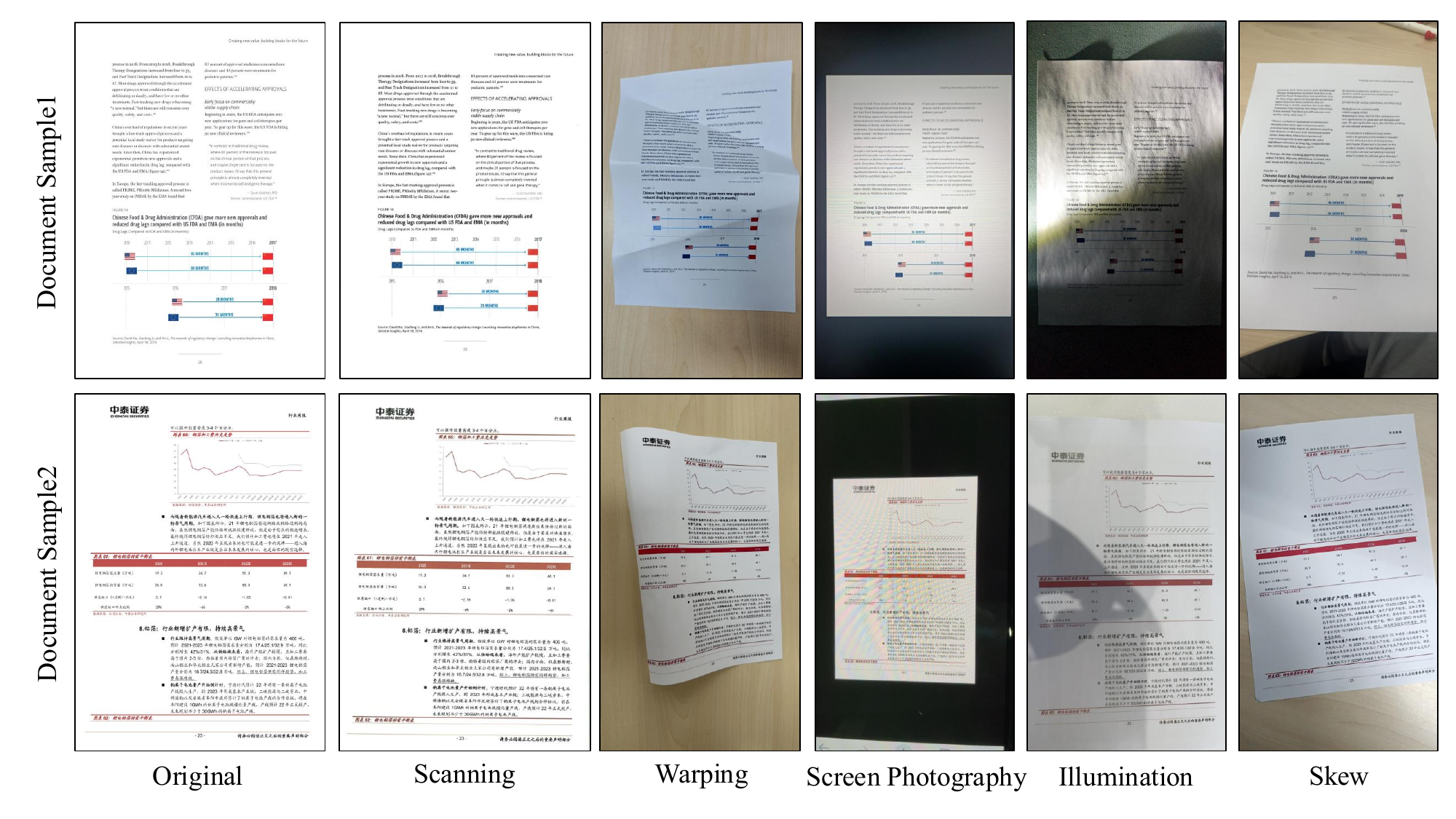} 

\caption{
    \centering
    Samples of Real5-OmniDocBench.
}
\label{fig:Real5-OmniDocBench_sample}
\end{figure}

Real5-OmniDocBench~\cite{zhou2026real5omnidocbenchfullscalephysicalreconstruction} is a brand-new benchmark oriented toward real-world scenarios, which we constructed based on the OmniDocBench v1.5 dataset. The dataset comprises five distinct scenarios: scanning, warping, screen photography, illumination, and Skew. Apart from the Scanning category, all images were manually acquired via handheld mobile devices to closely simulate real-world conditions. Each subset maintains a one-to-one correspondence with the original OmniDocBench, strictly adhering to its ground-truth annotations and evaluation protocols. Given its empirical and realistic nature, this dataset serves as a rigorous benchmark for assessing the robustness of document parsing models in practical applications. Figure  \ref{fig:Real5-OmniDocBench_sample} illustrates the visualization of representative samples from the proposed dataset.

As illustrated in Table \ref{tab:real5-omnidocbench-performance}, PaddleOCR-VL-1.5 demonstrates consistent superiority across all evaluated scenarios, setting a new SOTA record with an overall accuracy of 92.05\%. Despite its compact 0.9B parameter scale, the model significantly outperforms massive general VLMs, such as Qwen3-VL-235B and Gemini-3 Pro, highlighting its exceptional parameter efficiency for document-centric tasks. Notably, in the highly challenging Skewing category, PaddleOCR-VL-1.5 achieves an accuracy of 91.66\%, representing a 14.19\% absolute improvement over its predecessor. This substantial performance leap underscores its superior robustness against extreme geometric distortions and validates its reliability for complex document parsing in unconstrained environments. Detailed comparisons across sub-items including text, formulas, tables, and reading order can be found in Appendix~\ref{sec:Details of the Real5-OmniDocBench Benchmark}.

\begin{table}[H]
    \centering
    \resizebox{\textwidth}{!}{%
    \renewcommand{\arraystretch}{1.2}
    \begin{tabular}{l|ll|c|c c c c c}
        \toprule
        \textbf{Model Type} & \textbf{Methods} & \textbf{Parameters} & \textbf{Overall$\uparrow$} & \textbf{Scanning$\uparrow$} & \textbf{Warping$\uparrow$
} & \textbf{Screen Photography$\uparrow$} & \textbf{Illumination$\uparrow$} & \textbf{Skew$\uparrow$}\\    \midrule
        \multirow{2}{*}{\textbf{Pipeline Tools}} 
         & Maker-1.8.2~\cite{vik2024marker} & - & 60.10 & 70.27 & 58.98 & 63.65 & 66.31 &41.27 \\
         & PP-StructureV3~\cite{cui2025paddleocr} & - & 64.45 & 84.68 & 59.34 & 66.89 & 73.38 & 37.98 \\
         \midrule
         \multirow{5}{*}{\textbf{General VLMs}} 
         & GPT-5.2~\cite{gpt5_2} & - & 78.66 & 84.43 & 76.26 & 76.75 & 80.88 & 75.00 \\
         & Qwen2.5-VL-72B~\cite{bai2025qwen2} & 72B & 86.92 & 86.19 & 87.77 & 86.48 & 87.25 & 86.90 \\
         & Gemini-2.5 Pro~\cite{gemini25} & - & 88.21 & 89.25 & 87.63 & 87.11 & 87.97 & 89.07\\
         & Qwen3-VL-235B-A22B-Instruct~\cite{yang2025qwen3} & 235B & 88.904 & 89.43 & \cellcolor{cyan!15}\underline{89.99} & 89.27 & 89.27 & 86.56 \\
         & Gemini-3 Pro~\cite{gemini30} & - & \cellcolor{cyan!15}\underline{89.24} & 89.47 & 88.90 & 88.86 & 89.53 & \cellcolor{cyan!15}\underline{89.45} \\
         \midrule
         \multirow{10}{*}{\textbf{Specialized VLMs}}
         & Dolphin-1.5 ~\cite{feng2025dolphin} & 0.3B & 61.48 & 83.39 & 50.50 & 69.76 & 75.61 & 28.16\\
         & Dolphin~\cite{feng2025dolphin} & 0.3B & 61.78 & 72.16 & 60.35 & 64.29 & 67.29  & 44.83\\
         & Deepseek-OCR~\cite{wei2025deepseek} & 3B & 73.99 & 86.17 & 67.20 & 75.31 & 78.10 & 63.01\\
         & MinerU2-VLM~\cite{MinerU2} & 0.9B & 76.95 & 83.60 & 73.73 & 78.77 & 80.51 & 68.16\\
         & MonkeyOCR-pro-1.2B ~\cite{li2025monkeyocr} & 1.9B & 77.15 & 84.64 & 76.59 & 80.24 & 82.11 & 62.18\\
         & MonkeyOCR-3B ~\cite{li2025monkeyocr} & 3.7B & 78.29 & 84.65 & 77.27 & 80.71 & 83.16 & 65.67\\
         & MonkeyOCR-pro-3B ~\cite{li2025monkeyocr} & 3.7B & 79.49 & 86.94 & 78.90 & 82.44 & 84.71 & 64.47\\
         & Nanonets-OCR-s~\cite{Nanonets-OCR-S} & 3B & 84.19 & 85.52 & 83.56 & 84.86 & 85.01 & 81.98\\
         & PaddleOCR-VL~\cite{cui2025paddleocrvl} & 0.9B & 85.54 & \cellcolor{cyan!15}\underline{92.11} & 85.97 & 82.54 & \cellcolor{cyan!15}\underline{89.61} & 77.47\\
         & MinerU2.5~\cite{niu2025mineru2} & 1.2B & 85.61 & 90.06 & 83.76 & \cellcolor{cyan!15}\underline{89.41} & 89.57 & 75.24\\
         & dots.ocr~\cite{dotsocr} & 3B & 86.38 & 86.87 & 86.01 & 87.18 & 87.57 & 84.27\\
         & \textbf{PaddleOCR-VL-1.5} & 0.9B & \cellcolor{red!15}\textbf{92.05} & \cellcolor{red!15}\textbf{93.43} & \cellcolor{red!15}\textbf{91.25} & \cellcolor{red!15}\textbf{91.76} & \cellcolor{red!15}\textbf{92.16} & \cellcolor{red!15}\textbf{91.66}\\
        \bottomrule
    \end{tabular}%
    }
    \caption{Comprehensive evaluation of document parsing on Real5-OmniDocBench, Appendix \ref{sec:Details of the Real5-OmniDocBench Benchmark} provides more detailed metrics of this benchmark.}
    \label{tab:real5-omnidocbench-performance}
\end{table}

\subsection{New Capabilities}

\subsubsection{Text Spotting}
To thoroughly assess the model’s end-to-end text spotting capability (detection + recognition), we establish a comprehensive OCR benchmark spanning 10 key dimensions. In addition to standard settings such as common scenes (Common) and multilingual recognition (Japanese), the benchmark is designed to reflect practical deployment challenges by deliberately sampling more difficult cases, including degraded or low-quality images (Blur), highly variable handwriting in both Chinese and English (Handwrite\_ch/en), structured and layout-sensitive table content (Table), and culturally significant historical materials such as ancient documents and Traditional Chinese (Ancient). As summarized in Table~\ref{tab:spotting}, our model delivers the highest spotting accuracy across all 9 dimensions, consistently outperforming strong baselines and demonstrating robust generalization under diverse visual conditions and text styles. These results indicate that the proposed approach remains reliable not only in regular document scenarios but also in challenging, real-world settings that require precise localization and faithful transcription.
\begin{table}[htbp]
\centering
\resizebox{\textwidth}{!}{
\begin{tabular}{l|c|cccccccccc}
\toprule
\textbf{Dataset} 
& \textbf{Overall}
& \textbf{Ancient} 
& \textbf{Blur} 
& \textbf{Common} 
& \textbf{\makecell{Handwrite\\\_ch}} 
& \textbf{\makecell{Handwrite\\\_en}} 
& \textbf{\makecell{Printing\\\_ch}} 
& \textbf{\makecell{Printing\\\_en}} 
& \textbf{Table} 
& \textbf{Japanese} \\ 

\midrule

HunyuanOCR~\cite{hunyuanvisionteam2025hunyuanocrtechnicalreport}
& 0.6290
& 0.6164 & 0.6392 & 0.5222 & 0.7984 & 0.7665 & 0.6213 & 0.5956 & 0.4419  & 0.6593 \\

Rex-Omni~\cite{jiang2025detect} 
& 0.6682
& 0.4251 & 0.6936 & 0.6112 & 0.8147 & 0.7812 & 0.6961 & 0.6088 & 0.7185  & 0.6642\\

\textbf{PaddleOCR-VL-1.5} 
& \cellcolor{red!15}\textbf{0.8621}
& \cellcolor{red!15}\textbf{0.8523} 
& \cellcolor{red!15}\textbf{0.8422} 
& \cellcolor{red!15}\textbf{0.7713} 
& \cellcolor{red!15}\textbf{0.8952} 
& \cellcolor{red!15}\textbf{0.9163} 
& \cellcolor{red!15}\textbf{0.8669} 
& \cellcolor{red!15}\textbf{0.8689} 
& \cellcolor{red!15}\textbf{0.8993} 
& \cellcolor{red!15}\textbf{0.8461} \\ 

\bottomrule
\end{tabular}
}
\caption{Comparison of text spotting performance on the in-house benchmark. Overall denotes the average accuracy across all 9 evaluation dimensions.}
\label{tab:spotting}
\end{table}

\subsubsection{Seal Recognition}
\label{Seal Recognition}

To evaluate the effectiveness of our model in complex seal recognition tasks, we constructed a specialized benchmark comprising 300 high-quality images. This evaluation set covers a diverse range of seal shapes (e.g., circular, oval, and rectangular) and challenging real-world scenarios, such as overlapping text, low-contrast impressions, and distorted backgrounds. We employ the Normalized Edit Distance (NED) as the primary evaluation metric to assess recognition accuracy at the character level.

As illustrated in Table \ref{table:seal_results}, PaddleOCR-VL-1.5 demonstrates a clear advantage in seal recognition. Despite its compact size (0.9B parameters), it achieves an NED of 0.138, outperforming the 235B-parameter Qwen3-VL by a large margin (0.382). This highlights the model's effectiveness in handling specialized document elements.

\begin{table}[ht]
\centering
 \fontsize{8}{8}\selectfont
\renewcommand{\arraystretch}{1.2}

\begin{tabular}{l|c|c}
\toprule
\textbf{Model} & \textbf{Parameters} & \textbf{NED ($\downarrow$)} \\ 
\midrule
Qwen2.5-VL-72B~\cite{bai2025qwen2} & 72B & 0.396 \\
Qwen3-VL-235B-A22B-Instruct~\cite{yang2025qwen3} & 235B & 0.382 \\ 
\textbf{PaddleOCR-VL-1.5} & \cellcolor{red!15}\textbf{0.9B} & \cellcolor{red!15}\textbf{0.138} \\ 
\bottomrule
\end{tabular}
\caption{Comparison of seal recognition performance on in-house-seal benchmark.}
\label{table:seal_results}
\end{table}

\subsection{Inference Performance}

To speed up inference, we optimize the execution workflow of PaddleOCR-VL-1.5 by introducing an asynchronous, multi-threaded design, following the same strategy adopted in PaddleOCR-VL. The entire workflow is decomposed into three consecutive stages: input preparation (primarily converting PDF pages into images), layout analysis, and VLM inference. Each stage runs in its own dedicated thread, and intermediate results are exchanged between adjacent stages via queue-based buffers. This pipelined architecture enables concurrent execution across stages, thereby increasing parallelism and improving overall throughput. For the VLM inference stage in particular, mini-batches are formed dynamically: a batch is launched either when the queue size reaches a preset capacity or when the oldest queued item has waited longer than a specified time limit. This batching strategy makes it possible to group content blocks from multiple pages into a single inference call, which substantially increases parallel efficiency, especially when processing large collections of documents. In addition, we deploy PaddleOCR-VL-1.5-0.9B on high-performance inference and serving frameworks, i.e., FastDeploy~\cite{PaddlePaddle_FastDeploy}, vLLM~\cite{kwon2023efficient}, and SGLang~\cite{zheng2024sglang}. Key runtime parameters, including \texttt{max-num-batched-tokens} and \texttt{gpu-memory-utilization}, are carefully tuned to strike a balance between maximizing inference throughput and controlling GPU memory usage.


Table~\ref{tab:inference_performance} summarizes the end-to-end inference efficiency of different OCR methods under various deployment backends on the OmniDocBench v1.5 dataset. PaddleOCR-VL-1.5 achieves the best overall performance across all metrics. With the FastDeploy
 backend, it reaches 1.4335 pages/s and 2016.6 tokens/s on a single NVIDIA A100 GPU, surpassing its predecessor PaddleOCR-VL by 16.9\% and 18.6\%, respectively. These results verify that PaddleOCR-VL-1.5 provides state-of-the-art inference speed and throughput, making it well-suited for large-scale, real-world document understanding applications.

\begin{table}[H]
  \centering
  \fontsize{8}{8}\selectfont
  \renewcommand{\arraystretch}{1.2}
  \begin{tabular}{l|cccccc}
    \toprule
    \textbf{Methods} & \textbf{Backend} & \textbf{Total Time (s)$\downarrow$} & \textbf{Pages/s$\uparrow$} & \textbf{Tokens/s$\uparrow$} 
    \\
    \midrule
    MonkeyOCR-pro-1.2B~\cite{li2025monkeyocr}& vLLM (v0.10.2) & 2152.7 & 0.6292 & 949.8 & 
    \\
    dots.ocr~\cite{dotsocr} & vLLM (v0.14.0)& 3236.2 & 0.2791 & 374.3 & 
    \\
    MinerU2.5 (mineru=2.5.2)~\cite{niu2025mineru2} & vLLM (v0.10.2)& 1356.5 & 0.9984 & 1415.1 & 
    \\
    DeepSeek-OCR~\cite{wei2025deepseek} & vLLM (v0.8.5)& 2130.5 & 0.6358 & 897.4 & 
    \\
    PaddleOCR-VL & vLLM (v0.10.2) & 1325.5 & 1.0216 & 1419.9 & 
    \\
    PaddleOCR-VL & FastDeploy (v2.3) & \cellcolor{cyan!15}\underline{1104.5} & \cellcolor{cyan!15}\underline{1.2261} & \cellcolor{cyan!15}\underline{1700.5} & 
    \\
    PaddleOCR-VL-1.5 & vLLM (v0.10.2) & 1184.3 & 1.1433 & 1605.6 & 
    \\
    PaddleOCR-VL-1.5 & SGLang (v0.5.2)& 1342.0 & 1.0091 & 1418.9 & 
    \\
    PaddleOCR-VL-1.5 & FastDeploy (v2.3) & \cellcolor{red!15}\textbf{944.4} & \cellcolor{red!15}\textbf{1.4335} & \cellcolor{red!15}\textbf{2016.6} & 
    \\
    \bottomrule
  \end{tabular}
  \caption{
    End-to-End Inference Performance Comparison on OmniDocBench v1.5. PDF documents were processed in batches of 512 on a single NVIDIA A100 GPU. The reported end-to-end runtime includes both PDF rendering and Markdown generation. All methods rely on their built-in PDF parsing modules and default DPI settings to reflect out-of-the-box performance. Tokenization and special processing details follow the protocol introduced in \cite{cui2025paddleocrvl}.
  }
  \label{tab:inference_performance}
\end{table}

\section{Conclusion}

This work introduces PaddleOCR-VL-1.5, achieving a record SOTA accuracy of 94.5\% on OmniDocBench v1.5 and demonstrating superior general precision in document parsing. A key advancement of this version is its exceptional robustness in unconstrained real-world environments. The model effectively overcomes critical hurdles such as aggressive skewing, non-rigid page warping, and erratic lighting—scenarios where traditional solutions often fail. Furthermore, it expands its functional versatility with the integration of Seal Recognition and Text Spotting. By delivering a high-fidelity data foundation, PaddleOCR-VL-1.5 will significantly enhance the reliability and performance of downstream RAG systems and Large Language Model applications in complex, real-world deployment.
\bibliography{main}

\setcounter{figure}{0}
\makeatletter 
\renewcommand{\thefigure}{A\@arabic\c@figure}
\makeatother

\setcounter{table}{0}
\makeatletter 
\renewcommand{\thetable}{A\@arabic\c@table}
\makeatother

\clearpage 
\newpage
\appendix

\section*{Appendix}
\label{sec:appendix}

\section{Comparison of PaddleOCR-VL-1.5 and 1.0 Models}
\label{sec:Comparison of PaddleOCR-VL-1.5 and 1.0 Models}

\begin{table}[htbp]
\centering
\renewcommand{\arraystretch}{1.1} 
\small 

\begin{tabularx}{\textwidth}{l| c |c |c | X}
\toprule 
\textbf{Category} & \textbf{Capability Item} & \textbf{V1} & \textbf{V1.5} & \textbf{V1.5 Description} \\ 
\midrule

\multirow{13}{*}{Fundamental} 
 & \multirow{3}{*}{Layout Analysis} & \multirow{3}{*}{\fstar\fstar\estar\estar\estar }&  \multirow{3}{*}{\fstar\fstar\fstar\fstar\estar} & Improved stability for warped/skewed scenes; added CAD and comics. \\
 \cline{2-5}
 & \multirow{3}{*}{Text Recognition} & \multirow{3}{*}{\fstar\fstar\fstar\fstar\estar} & \multirow{3}{*}{\fstar\fstar\fstar\fstar\fstar} & Gains in vertical text, special characters, and emphasis marks. \\
  \cline{2-5}
 & \multirow{2}{*}{Table Recognition} & \multirow{2}{*}{\fstar\fstar\fstar\estar\estar} & \multirow{2}{*}{\fstar\fstar\fstar\fstar\estar} & Improvements for borderless tables and invoices. \\
  \cline{2-5}
 & \multirow{2}{*}{Formula Recognition} & \multirow{2}{*}{\fstar\fstar\fstar\estar\estar} & \multirow{2}{*}{\fstar\fstar\fstar\fstar\estar} & Better in skewed formulas and illumination scenarios. \\
  \cline{2-5}
 & \multirow{2}{*}{Chart Recognition }& \multirow{2}{*}{\fstar\fstar\fstar\fstar\estar} & \multirow{2}{*}{\fstar\fstar\fstar\fstar\estar} & Capability remains consistent with the previous version. \\
  \cline{2-5}
 & \multirow{2}{*}{Reading Order} & \multirow{2}{*}{\fstar\fstar\fstar\estar\estar} & \multirow{2}{*}{\fstar\fstar\fstar\fstar\estar} & Significant boost for irregular layouts. \\
\midrule

\multirow{11}{*}{Adaptability} 
 & \multirow{2}{*}{Skewed Docs} & \multirow{2}{*}{\fstar\estar\estar\estar\estar} & \multirow{2}{*}{\fstar\fstar\fstar\fstar\fstar} & Dramatic improvement for high-angle tilted documents. \\
   \cline{2-5}
 & \multirow{2}{*}{Scanned Docs} & \multirow{2}{*}{\fstar\fstar\fstar\fstar\estar} & \multirow{2}{*}{\fstar\fstar\fstar\fstar\fstar} & Stability for low-quality scans is significantly enhanced. \\
   \cline{2-5}
 & \multirow{3}{*}{Warped Docs} & \multirow{3}{*}{\fstar\fstar\estar\estar\estar} & \multirow{3}{*}{\fstar\fstar\fstar\fstar\fstar} & Supports complex physical deformation and folded paper. \\
   \cline{2-5}
 & \multirow{3}{*}{Screen Photo} & \multirow{3}{*}{\fstar\fstar\estar\estar\estar} & \multirow{3}{*}{\fstar\fstar\fstar\fstar\fstar} & Suppresses interference from reflections and Moiré patterns. \\
   \cline{2-5}
 & \multirow{2}{*}{Illumination} & \multirow{2}{*}{\fstar\fstar\fstar\fstar\estar} & \multirow{2}{*}{\fstar\fstar\fstar\fstar\fstar} & Superior performance in uneven or weak lighting. \\

\midrule 
\multirow{8}{*}{New Features} 
 & \multirow{2}{*}{Seal Recognition} & \multirow{2}{*}{\estar\estar\estar\estar\estar} & \multirow{2}{*}{\fstar\fstar\fstar\fstar\estar} & Recognition of various official seals and stamps. \\
   \cline{2-5}
 & \multirow{2}{*}{Text Spotting} & \multirow{2}{*}{\estar\estar\estar\estar\estar} & \multirow{2}{*}{\fstar\fstar\fstar\fstar\estar} & Localization and recognition of multiple character sets. \\  \cline{2-5}
 &\multirow{2}{*}{ Cross-page Table Merginig} & \multirow{2}{*}{\estar\estar\estar\estar\estar} & \multirow{2}{*}{\fstar\fstar\fstar\fstar\estar} & Merges split tables while maintaining consistency. \\  \cline{2-5}
 & \multirow{2}{*}{Heading Hierarchy} & \multirow{2}{*}{\estar\estar\estar\estar\estar} & \multirow{2}{*}{\fstar\fstar\fstar\estar\estar} & Title hierarchy recognition across multi-page documents. \\[1em]
\bottomrule 
\end{tabularx}

\caption{Comprehensive functional evolution and robustness comparison between PaddleOCR-VL and PaddleOCR-VL-1.5. The star ratings only indicate the relative performance of the two versions and do not represent their absolute accuracy.}
\label{tab:capability_evolution}
\end{table}

\clearpage 
\newpage

\section{Details of the Real5-OmniDocBench Benchmark}
\label{sec:Details of the Real5-OmniDocBench Benchmark}
Real5-OmniDocBench\footnote{https://huggingface.co/datasets/PaddlePaddle/Real5-OmniDocBench} is a brand-new benchmark oriented toward real-world scenarios, which we constructed based on the OmniDocBench v1.5~\cite{ouyang2025omnidocbench} dataset. PaddleOCR-VL-1.5 achieves state-of-the-art (SOTA) results across all sub-scenarios within Real5-OmniDocBench, demonstrating its robust parsing capabilities for real-world documents. A detailed comparison of PaddleOCR-VL-1.5 against other advanced document parsing models across various metrics on this dataset is provided in this Appendix.

As shown in Table~\ref{tab:ocr_performance_scaning}, under the scaning scenario, PaddleOCR-VL-1.5 achieves state-of-the-art performance across all key metrics, consistently outperforming existing pipeline tools, general vision-language models, and specialized document parsing models. Compared to its predecessor, PaddleOCR-VL, the new version maintains a compact parameter size of 0.9B while raising the overall score from 92.11\% to a leading 93.43\%. Notably, PaddleOCR-VL-1.5 sets new records in all sub-tasks within this scenario, including a Formula-CDM score of 93.04\% and a Table-TEDS score of 90.97\%, both significantly surpassing larger models such as Qwen3-VL-235B and Gemini-3 Pro. Additionally, the model achieves exceptionally low Text-Edit Distance (0.037) and Reading Order score (0.045), further demonstrating its high accuracy in text recognition, formula extraction, and complex table structure analysis. Overall, PaddleOCR-VL-1.5 delivers a new breakthrough in the Real5-OmniDocBench-scaning scenario.

\begin{table}[H]
    \centering
    \resizebox{\textwidth}{!}{%
    \renewcommand{\arraystretch}{1.2}
    \begin{tabular}{l|ll|c|c c c c}
        \toprule
        \textbf{Model Type} & \textbf{Methods} & \textbf{Parameters} & \textbf{Overall$\uparrow$} & \textbf{Text\textsuperscript{Edit}$\downarrow$} & \textbf{Formula\textsuperscript{CDM}$\uparrow$} & \textbf{Table\textsuperscript{TEDS}$\uparrow$} & \textbf{Reading Order\textsuperscript{Edit}$\downarrow$} \\    \midrule
        \textbf{Pipeline Tools} 
        & Maker-1.8.2~\cite{vik2024marker} & - & 70.27 & 0.223 & 77.03 & 56.05  & 0.238 \\
        & PP-StructureV3~\cite{cui2025paddleocr} & - & 84.68 & 0.094 & 84.34 & 79.06  & 0.092 \\
        \midrule
        \multirow{5}{*}{\textbf{General VLMs}} 
        & GPT-5.2~\cite{gpt5_2} & - & 84.43 & 0.142 & 85.68 & 81.78  & 0.109 \\
        & Qwen2.5-VL-72B~\cite{bai2025qwen2} & 72B & 86.19 & 0.110 & 86.14 & 83.41  & 0.114 \\
        
         & Gemini-2.5 Pro~\cite{gemini25} & - & 89.25 & 0.073 & 87.44 & 87.62 & 0.098 \\
         & Qwen3-VL-235B-A22B-Instruct~\cite{bai2025qwen2} & 235B & 89.43 & 0.059 & 89.01 & 85.19 & 0.066 \\
         & Gemini-3 Pro~\cite{gemini30} & - & 89.47 & 0.071 & 88.16 & 87.37 & 0.078 \\
        \midrule
        \multirow{10}{*}{\textbf{Specialized VLMs}} & Dolphin~\cite{feng2025dolphin} & 322M & 72.16 & 0.154 & 64.58 & 67.27 & 0.130 \\
         & Dolphin-1.5 ~\cite{feng2025dolphin} & 0.3B & 83.39 & 0.097 & 76.25 & 83.65 & 0.090 \\
         & MinerU2-VLM~\cite{MinerU2} & 0.9B & 83.60 & 0.094 & 79.76 & 80.44 & 0.091 \\
         & MonkeyOCR-pro-1.2B ~\cite{li2025monkeyocr} & 1.9B & 84.64 & 0.123 & 84.17 & 82.13 & 0.145 \\
         & MonkeyOCR-3B ~\cite{li2025monkeyocr} & 3.7B & 84.65 & 0.100 & 84.16 & 79.81 & 0.143 \\
         & Nanonets-OCR-s~\cite{Nanonets-OCR-S} & 3B & 85.52 & 0.106 & 88.09 & 79.11 & 0.106 \\
         & Deepseek-OCR~\cite{wei2025deepseek} & 3B & 86.17 & 0.078 & 83.59 & 82.69 & 0.085 \\
         & dots.ocr~\cite{dotsocr} & 3B & 86.87 & 0.083 & 83.27 & 85.68 & 0.081 \\
         & MonkeyOCR-pro-3B ~\cite{li2025monkeyocr} & 3.7B & 86.94 & 0.103 & 86.29 & 84.86 & 0.141 \\
         & MinerU2.5~\cite{niu2025mineru2} & 1.2B & 90.06 & 0.052 & 88.22 & 87.16 & 0.050 \\
         & PaddleOCR-VL~\cite{cui2025paddleocrvl} & 0.9B & \cellcolor{cyan!15}\underline{92.11} & \cellcolor{cyan!15}\underline{0.039} & \cellcolor{cyan!15}\underline{90.35} & \cellcolor{cyan!15}\underline{89.90} & \cellcolor{cyan!15}\underline{0.048} \\ 
         & \textbf{PaddleOCR-VL-1.5} & 0.9B & \cellcolor{red!15}\textbf{93.43} & \cellcolor{red!15}\textbf{0.037} & \cellcolor{red!15}\textbf{93.04} & \cellcolor{red!15}\textbf{90.97} & \cellcolor{red!15}\textbf{0.045} \\
        \bottomrule
    \end{tabular}%
    }
    \caption{Comprehensive evaluation of document parsing on Real5-OmniDocBench-scaning}
    \label{tab:ocr_performance_scaning}
\end{table}

As shown in Table~\ref{tab:ocr_performance_warping}, PaddleOCR-VL-1.5 exhibits notable robustness in the warping scenario, achieving an overall score of 91.25\%, which is higher than the larger Qwen3-VL-235B model (89.99\%). Its Formula-CDM score of 90.94\% and Table-TEDS score of 88.10\% indicate a strong ability to preserve document structure under significant geometric distortion.

\begin{table}[H]
    \centering
    \resizebox{\textwidth}{!}{%
    \renewcommand{\arraystretch}{1.2}
    \begin{tabular}{l|ll|c|c c c c}
        \toprule
        \textbf{Model Type} & \textbf{Methods} & \textbf{Parameters} & \textbf{Overall$\uparrow$} & \textbf{Text\textsuperscript{Edit}$\downarrow$} & \textbf{Formula\textsuperscript{CDM}$\uparrow$} & \textbf{Table\textsuperscript{TEDS}$\uparrow$} & \textbf{Reading Order\textsuperscript{Edit}$\downarrow$} \\    \midrule
        \textbf{Pipeline Tools} 
        & Maker-1.8.2~\cite{vik2024marker} & - & 58.98 & 0.349 & 72.71 & 39.08  & 0.390 \\
        & PP-StructureV3~\cite{cui2025paddleocr} & - & 59.34 & 0.376 & 68.22 & 47.40  & 0.261 \\
        \midrule
        \multirow{5}{*}{\textbf{General VLMs}} 
        & GPT-5.2~\cite{gpt5_2} & - & 76.26 & 0.239 & 80.90 & 71.80  & 0.165 \\
        & Gemini-2.5 Pro~\cite{gemini25} & - & 87.63 & 0.092 & 86.50 & 85.59 & 0.109 \\
        & Qwen2.5-VL-72B~\cite{bai2025qwen2} & 72B & 87.77 & 0.086 & 88.85 & 83.06  & 0.102 \\
         & Gemini-3 Pro~\cite{gemini30} & - & 88.90 & 0.086 & 88.10 & \cellcolor{cyan!15}\underline{87.20} & 0.087 \\
         & Qwen3-VL-235B-A22B-Instruct~\cite{bai2025qwen2} & 235B & \cellcolor{cyan!15}\underline{89.99} & \cellcolor{red!15}\textbf{0.051} & \cellcolor{cyan!15}\underline{89.06} & 85.95 &  \cellcolor{cyan!15}\underline{0.064} \\
        \midrule
        \multirow{10}{*}{\textbf{Specialized VLMs}} & Dolphin-1.5 ~\cite{feng2025dolphin} & 0.3B & 50.50 & 0.383 & 47.24 & 42.52 & 0.309 \\
        & Dolphin~\cite{feng2025dolphin} & 322M & 60.35 & 0.316 & 61.06 & 51.58 & 0.247 \\
        & Deepseek-OCR~\cite{wei2025deepseek} & 3B & 67.20 & 0.328 & 73.59 & 60.80 & 0.226 \\
         & MinerU2-VLM~\cite{MinerU2} & 0.9B & 73.73 & 0.202 & 77.72 & 63.65 & 0.173 \\
         & MonkeyOCR-pro-1.2B ~\cite{li2025monkeyocr}& 1.9B & 76.59 & 0.196 & 78.85 & 70.52 & 0.221 \\
         & MonkeyOCR-3B ~\cite{li2025monkeyocr}& 3.7B & 77.27 & 0.164 & 79.08 & 69.18 & 0.211 \\
         & MonkeyOCR-pro-3B ~\cite{li2025monkeyocr}& 3.7B & 78.90 & 0.168 & 79.55 & 73.94 & 0.212 \\
         & Nanonets-OCR-s~\cite{Nanonets-OCR-S} & 3B & 83.56 & 0.121 & 86.24 & 76.57 & 0.124 \\
         & MinerU2.5~\cite{niu2025mineru2} & 1.2B & 83.76 & 0.154 & 85.92 & 80.71 & 0.104 \\
         & PaddleOCR-VL ~\cite{cui2025paddleocrvl}& 0.9B & 85.97 & 0.093 & 85.45 & 81.77 & 0.092 \\ 
         & dots.ocr~\cite{dotsocr} & 3B & 86.01 & 0.087 & 85.03 & 81.74 & 0.093 \\
         & \textbf{PaddleOCR-VL-1.5} & 0.9B & \cellcolor{red!15}\textbf{91.25} & \cellcolor{cyan!15}\underline{0.053} & \cellcolor{red!15}\textbf{90.94} & \cellcolor{red!15}\textbf{88.10} & \cellcolor{red!15}\textbf{0.063} \\
        \bottomrule
    \end{tabular}%
    }
    \caption{Comprehensive evaluation of document parsing on Real5-OmniDocBench-warping.}
    \label{tab:ocr_performance_warping}
\end{table}

In the screen photography scenario presented in Table~\ref{tab:ocr_performance_screen_photography}, PaddleOCR-VL-1.5 attains an overall score of 91.76\%, demonstrating competitive performance among specialized vision-language models. The model achieves a Formula-CDM score of 90.88\%, outperforming MinerU2.5 (87.55\%) and dots.ocr (85.34\%), and shows effective handling of Moire patterns and reflections typically encountered in screen-captured documents.

\begin{table}[H]
    \centering
    \resizebox{\textwidth}{!}{%
    \renewcommand{\arraystretch}{1.2}
    \begin{tabular}{l|ll|c|c c c c}
        \toprule
        \textbf{Model Type} & \textbf{Methods} & \textbf{Parameters} & \textbf{Overall$\uparrow$} & \textbf{Text\textsuperscript{Edit}$\downarrow$} & \textbf{Formula\textsuperscript{CDM}$\uparrow$} & \textbf{Table\textsuperscript{TEDS}$\uparrow$} & \textbf{Reading Order\textsuperscript{Edit}$\downarrow$} \\    \midrule
        \textbf{Pipeline Tools} 
        & Maker-1.8.2~\cite{vik2024marker} & - & 63.65 & 0.290 & 72.73 & 47.21  & 0.325 \\
        & PP-StructureV3~\cite{cui2025paddleocr} & - & 66.89 & 0.204 & 73.26 & 47.82  & 0.165 \\
        \midrule
        \multirow{5}{*}{\textbf{General VLMs}} 
        & GPT-5.2~\cite{gpt5_2} & - & 76.75 & 0.208 & 79.27 & 71.73  & 0.148 \\
        & Qwen2.5-VL-72B~\cite{bai2025qwen2} & 72B & 86.48 & 0.100 & 87.46 & 82.00  & 0.102 \\
        & Gemini-2.5 Pro~\cite{gemini25} & - & 87.11 & 0.103 & 85.30 & 86.31 & 0.117 \\
         & Gemini-3 Pro~\cite{gemini30} & - & 88.86 & 0.084 & 87.33 & \cellcolor{cyan!15}\underline{87.65} & 0.087 \\
         & Qwen3-VL-235B-A22B-Instruct~\cite{bai2025qwen2} & 235B & \cellcolor{cyan!15}\underline{89.27} & \cellcolor{cyan!15}\underline{0.068} & \cellcolor{cyan!15}\underline{88.72} & 85.85 &  0.071 \\
        \midrule
        \multirow{10}{*}{\textbf{Specialized VLMs}} & Dolphin~\cite{feng2025dolphin} & 322M & 64.29 & 0.232 & 58.66 & 57.38 & 0.195 \\
        & Dolphin-1.5 ~\cite{feng2025dolphin} & 0.3B & 69.76 & 0.205 & 61.80 & 68.00 & 0.177 \\
        & Deepseek-OCR~\cite{wei2025deepseek} & 3B & 75.31 & 0.220 & 77.68 & 70.26 & 0.169 \\
         & MinerU2-VLM~\cite{MinerU2} & 0.9B & 78.77 & 0.139 & 79.02 & 71.17 & 0.123 \\
         & MonkeyOCR-pro-1.2B ~\cite{li2025monkeyocr}& 1.9B & 80.24 & 0.148 & 80.78 & 74.74 & 0.179 \\
         & MonkeyOCR-3B ~\cite{li2025monkeyocr}& 3.7B & 80.71 & 0.122 & 81.33 & 73.04 & 0.177 \\
         & MonkeyOCR-pro-3B ~\cite{li2025monkeyocr}& 3.7B & 82.44 & 0.124 & 81.55 & 78.13 & 0.177 \\
         & PaddleOCR-VL ~\cite{cui2025paddleocrvl}& 0.9B & 82.54 & 0.103 & 83.58 & 74.36 & 0.107 \\
         & Nanonets-OCR-s~\cite{Nanonets-OCR-S} & 3B & 84.86 & 0.112 & 86.65 & 79.09 & 0.117 \\
         & dots.ocr~\cite{dotsocr} & 3B & 87.18 & 0.081 & 85.34 & 84.26 & 0.079 \\
         & MinerU2.5~\cite{niu2025mineru2} & 1.2B & 89.41 & 0.062 & 87.55 & 86.83 & \cellcolor{red!15}\textbf{0.053} \\ 
         & \textbf{PaddleOCR-VL-1.5} & 0.9B & \cellcolor{red!15}\textbf{91.76} & \cellcolor{red!15}\textbf{0.050} & \cellcolor{red!15}\textbf{90.88} & \cellcolor{red!15}\textbf{89.38} & \cellcolor{cyan!15}\underline{0.059} \\
        \bottomrule
    \end{tabular}%
    }
    \caption{Comprehensive evaluation of document parsing on Real5-OmniDocBench-screen-photography.}
    \label{tab:ocr_performance_screen_photography}
\end{table}

Table~\ref{tab:ocr_performance_illumination} evaluates performance under illumination variations, where PaddleOCR-VL-1.5 reaches an overall score of 92.16\%. This result not only marks a significant improvement over the previous PaddleOCR-VL (89.61\%) but also surpasses top-tier general VLMs such as Gemini-3 Pro (89.53\%). The model’s Formula-CDM score of 91.80\% and Table-TEDS of 89.33\% underscore its high sensitivity and accuracy in low-contrast or unevenly lit environments.

\begin{table}[H]
    \centering
    \resizebox{\textwidth}{!}{%
    \renewcommand{\arraystretch}{1.2}
    \begin{tabular}{l|ll|c|c c c c}
        \toprule
        \textbf{Model Type} & \textbf{Methods} & \textbf{Parameters} & \textbf{Overall$\uparrow$} & \textbf{Text\textsuperscript{Edit}$\downarrow$} & \textbf{Formula\textsuperscript{CDM}$\uparrow$} & \textbf{Table\textsuperscript{TEDS}$\uparrow$} & \textbf{Reading Order\textsuperscript{Edit}$\downarrow$} \\    \midrule
        \textbf{Pipeline Tools} 
        & Maker-1.8.2~\cite{vik2024marker} & - & 66.31 & 0.259 & 74.80 & 50.03  & 0.337 \\
        & PP-StructureV3~\cite{cui2025paddleocr} & - & 73.38 & 0.158 & 77.75 & 58.19  & 0.126 \\
        \midrule
        \multirow{5}{*}{\textbf{General VLMs}} 
        & GPT-5.2~\cite{gpt5_2} & - & 80.88 & 0.191 & 84.41 & 77.37  & 0.134 \\
        & Qwen2.5-VL-72B~\cite{bai2025qwen2} & 72B & 87.25 & 0.087 & 86.44 & 84.03  & 0.097 \\
        & Gemini-2.5 Pro~\cite{gemini25} & - & 87.97 & 0.083 & 86.13 & 86.11 & 0.103 \\
         & Qwen3-VL-235B-A22B-Instruct~\cite{bai2025qwen2} & 235B & 89.27 & 0.060 & 87.81 & 86.05 &  0.070 \\
         & Gemini-3 Pro~\cite{gemini30} & - & 89.53 & 0.073 & 87.78 & 88.14 & 0.080 \\
        \midrule
        \multirow{10}{*}{\textbf{Specialized VLMs}} & Dolphin~\cite{feng2025dolphin} & 322M & 67.29 & 0.197 & 61.42 & 60.10 & 0.173 \\
        & Dolphin-1.5 ~\cite{feng2025dolphin} & 0.3B & 75.61 & 0.159 & 70.04 & 72.69 & 0.133 \\
        & Deepseek-OCR~\cite{wei2025deepseek} & 3B & 78.10 & 0.192 & 81.71 & 71.81 & 0.156 \\
         & MinerU2-VLM~\cite{MinerU2} & 0.9B & 80.51 & 0.135 & 80.72 & 74.29 & 0.123 \\
         & MonkeyOCR-pro-1.2B ~\cite{li2025monkeyocr}& 1.9B & 82.11 & 0.144 & 82.07 & 78.67 & 0.172 \\
         & MonkeyOCR-3B ~\cite{li2025monkeyocr}& 3.7B & 83.16 & 0.118 & 83.63 & 77.62 & 0.168 \\
         & MonkeyOCR-pro-3B ~\cite{li2025monkeyocr}& 3.7B & 84.71 & 0.120 & 84.13 & 82.02 & 0.171 \\
         & Nanonets-OCR-s~\cite{Nanonets-OCR-S} & 3B & 85.01 & 0.099 & 87.94 & 76.96 & 0.112 \\
         & dots.ocr~\cite{dotsocr} & 3B & 87.57 & 0.068 & 85.07 & 84.44 & 0.076 \\
         & MinerU2.5~\cite{niu2025mineru2} & 1.2B & 89.57 & 0.065 & \cellcolor{cyan!15}\underline{88.36} & 86.87 & 0.062 \\
         & PaddleOCR-VL ~\cite{cui2025paddleocrvl}& 0.9B & \cellcolor{cyan!15}\underline{89.61} & \cellcolor{cyan!15}\underline{0.049} & 86.66 & \cellcolor{cyan!15}\underline{87.02} & \cellcolor{cyan!15}\underline{0.055} \\ 
         & \textbf{PaddleOCR-VL-1.5} & 0.9B & \cellcolor{red!15}\textbf{92.16} & \cellcolor{red!15}\textbf{0.046} & \cellcolor{red!15}\textbf{91.80} & \cellcolor{red!15}\textbf{89.33} & \cellcolor{red!15}\textbf{0.051} \\
        \bottomrule
    \end{tabular}%
    }
    \caption{Comprehensive evaluation of document parsing on Real5-OmniDocBench-illumination.}
    \label{tab:ocr_performance_illumination}
\end{table}

Under the challenging skew detailed in Table~\ref{tab:ocr_performance_skew}, PaddleOCR-VL-1.5 maintains its dominance with an overall score of 91.66\%, once again outperforming general VLMs including Gemini-3 Pro (89.45\%). It particularly excels in complex structural recovery, evidenced by a Table-TEDS score of 91.00\% and a Text-Edit distance reduced to 0.047, demonstrating its superior ability to rectify and parse slanted document layouts.

\begin{table}[H]
    \centering
    \resizebox{\textwidth}{!}{%
    \renewcommand{\arraystretch}{1.2}
    \begin{tabular}{l|ll|c|c c c c}
        \toprule
        \textbf{Model Type} & \textbf{Methods} & \textbf{Parameters} & \textbf{Overall$\uparrow$} & \textbf{Text\textsuperscript{Edit}$\downarrow$} & \textbf{Formula\textsuperscript{CDM}$\uparrow$} & \textbf{Table\textsuperscript{TEDS}$\uparrow$} & \textbf{Reading Order\textsuperscript{Edit}$\downarrow$} \\    \midrule
        \textbf{Pipeline Tools} 
        & PP-StructureV3~\cite{cui2025paddleocr} & - & 37.98 & 0.557 & 44.37 & 25.27  & 0.417 \\
        & Maker-1.8.2~\cite{vik2024marker} & - & 41.27 & 0.536 & 60.16 & 17.23  & 0.543 \\
        \midrule
        \multirow{5}{*}{\textbf{General VLMs}} 
        & GPT-5.2~\cite{gpt5_2} & - & 75.00 & 0.257 & 80.27 & 70.47  & 0.167 \\
        & Qwen3-VL-235B-A22B-Instruct~\cite{bai2025qwen2} & 235B & 86.56 & \cellcolor{cyan!15}\underline{0.077} & 83.96 & 83.41 &  \cellcolor{cyan!15}\underline{0.091} \\
        & Qwen2.5-VL-72B~\cite{bai2025qwen2} & 72B & 86.90 & \cellcolor{cyan!15}\underline{0.077} & 87.26 & 81.14  & \cellcolor{cyan!15}\underline{0.091} \\
         & Gemini-2.5 Pro~\cite{gemini25} & - & 89.07 & \cellcolor{cyan!15}\underline{0.077} & 87.89 & 86.99 & 0.104 \\
         & Gemini-3 Pro~\cite{gemini30} & - & \cellcolor{cyan!15}\underline{89.45} & 0.080 & \cellcolor{cyan!15}\underline{88.33} & \cellcolor{cyan!15}\underline{88.06} & 0.092 \\
        \midrule
        \multirow{10}{*}{\textbf{Specialized VLMs}} & Dolphin-1.5 ~\cite{feng2025dolphin} & 0.3B & 28.16 & 0.553 & 25.60 & 14.18 & 0.419 \\
        & Dolphin~\cite{feng2025dolphin} & 322M & 44.83 & 0.500 & 51.34 & 33.22 & 0.321 \\
        & MonkeyOCR-pro-1.2B ~\cite{li2025monkeyocr}& 1.9B & 62.18 & 0.292 & 66.25 & 49.46 & 0.317 \\
        & Deepseek-OCR~\cite{wei2025deepseek} & 3B & 63.01 & 0.327 & 73.27 & 48.48 & 0.231 \\
        & MonkeyOCR-pro-3B ~\cite{li2025monkeyocr}& 3.7B & 64.47 & 0.251 & 69.06 & 49.42 & 0.301 \\
        & MonkeyOCR-3B ~\cite{li2025monkeyocr}& 3.7B & 65.67 & 0.248 & 69.23 & 52.59 & 0.300 \\
         & MinerU2-VLM~\cite{MinerU2} & 0.9B & 68.16 & 0.230 & 74.45 & 53.07 & 0.191 \\
         & MinerU2.5~\cite{niu2025mineru2} & 1.2B & 75.24 & 0.305 & 81.78 & 74.39 & 0.151 \\
         & PaddleOCR-VL ~\cite{cui2025paddleocrvl}& 0.9B & 77.47 & 0.192 & 78.81 & 72.83 & 0.193 \\
         & Nanonets-OCR-s~\cite{Nanonets-OCR-S} & 3B & 81.98 & 0.121 & 85.78 & 72.22 & 0.133 \\
         & dots.ocr~\cite{dotsocr} & 3B & 84.27 & 0.087 & 85.73 & 75.74 & 0.094 \\
         & \textbf{PaddleOCR-VL-1.5} & 0.9B & \cellcolor{red!15}\textbf{91.66} & \cellcolor{red!15}\textbf{0.047} & \cellcolor{red!15}\textbf{91.00} & \cellcolor{red!15}\textbf{88.69} & \cellcolor{red!15}\textbf{0.061} \\
        \bottomrule
    \end{tabular}%
    }
    \caption{Comprehensive evaluation of document parsing on Real5-OmniDocBench-skewing variation.}
    \label{tab:ocr_performance_skew}
\end{table}

\clearpage 
\newpage

\section{Supported Languages}

PaddleOCR-VL-1.5 supports a total of 111 languages. Compared to PaddleOCR-VL, PaddleOCR-VL-1.5 adds recognition capabilities for China's Tibetan script and Bengali. Table~\ref{tab:supported_language} lists the correspondence between each language category and the specific supported languages/scripts.

\begin{table}[ht]
\centering
\begin{tabular}{>{\centering\arraybackslash}m{0.35\textwidth}|>{\centering\arraybackslash}m{0.65\textwidth}}
\toprule
\textbf{Language Category} & \textbf{Specific Languages} \\
\midrule
Chinese & Chinese \\
\midrule
English & English \\
\midrule
Korean & Korean \\
\midrule
Japanese & Japanese \\
\midrule
Thai & Thai\\
\midrule
Greek & Greek \\
\midrule
Tamil & Tamil \\
\midrule
Telugu & Telugu \\
\midrule
\cellcolor{red!15}\textbf{Bengali*} & \cellcolor{red!15}\textbf{Bengali*} \\
\midrule
\cellcolor{red!15}\textbf{China's Tibetan script*} & \cellcolor{red!15}\textbf{China's Tibetan script*} \\
\midrule
Arabic & Arabic, Persian, Uyghur, Urdu, Pashto, Kurdish, Sindhi, Balochi \\
\midrule
Latin & French, German, Afrikaans, Italian, Spanish, Bosnian, Portuguese, Czech, Welsh, Danish, Estonian, Irish, Croatian, Uzbek, Hungarian, Serbian (Latin), Indonesian, Occitan, Icelandic, Lithuanian, Maori, Malay, Dutch, Norwegian, Polish, Slovak, Slovenian, Albanian, Swedish, Swahili, Tagalog, Turkish, Latin, Azerbaijani, Kurdish, Latvian, Maltese, Pali, Romanian, Vietnamese, Finnish, Basque, Galician, Luxembourgish, Romansh, Catalan, Quechua \\
\midrule
Cyrillic & Russian, Belarusian, Ukrainian, Serbian (Cyrillic), Bulgarian, Mongolian, Abkhazian, Adyghe, Kabardian, Avar, Dargin, Ingush, Chechen, Lak, Lezgin, Tabasaran, Kazakh, Kyrgyz, Tajik, Macedonian, Tatar, Chuvash, Bashkir, Malian, Moldovan, Udmurt, Komi, Ossetian, Buryat, Kalmyk, Tuvan, Sakha, Karakalpak \\
\midrule
Devanagari & Hindi, Marathi, Nepali, Bihari, Maithili, Angika, Bhojpuri, Magahi, Santali, Newari, Konkani, Sanskrit, Haryanvi \\
\bottomrule
\end{tabular}
   \caption{Supported Languages/scipts (*indicates newly added languages/scipts )}
       \label{tab:supported_language}
\end{table}

\clearpage 
\newpage

\section{Inference Performance on Different Hardware Configurations}

We evaluated the inference throughput and latency of PaddleOCR-VL-1.5 across multiple hardware configurations, with the detailed results presented in Table~\ref{tab:inference_performance_hardware}. In our experiments, all PDF pages are rendered at 72 DPI, which provides a good trade-off between memory efficiency and the visual fidelity required for reliable OCR. We note that the experiments on different hardware platforms were conducted without extensive parameter tuning or system-level optimization; therefore, the reported performance figures should be regarded as conservative and still leave room for further improvement. All models were evaluated using three deployment backends, namely FastDeploy v2.3.0, vLLM v0.10.2, and SGLang v0.5.2. Across these frameworks, PaddleOCR-VL-1.5 consistently delivers high and stable inference efficiency, demonstrating strong generalization across diverse hardware configurations and execution engines, as well as good compatibility with heterogeneous computing environments.

\begin{table}[H]
  \centering

  \renewcommand{\arraystretch}{1.2}
  \begin{tabular}{l|c|ccccc}
    \toprule
    \textbf{Hardware} & \textbf{Backend} & \textbf{Total Time (s)$\downarrow$} & \textbf{Pages/s$\uparrow$} & \textbf{Tokens/s$\uparrow$} & \textbf{Avg. VRAM Usage (GB)$\downarrow$} \\
    \midrule
    \multirow{3}{*}{H800} & FastDeploy & 556.4 & 2.4320 & 3404.5 & 64.8 \\ & vLLM & 761.8 & 1.7772 & 2488.0 & 46.2 \\
                          & SGLang & 868.5 & 1.5589 & 2185.2 & 48.9 \\
    \midrule
    \multirow{3}{*}{A100} & FastDeploy & 671.3 & 2.0160 & 2826.0 & 62.1 \\ & vLLM & 981.4 & 1.3797 & 1926.1 & 43.5 \\
                          & SGLang & 1100.9 & 1.2301 & 1722.5 & 48.9 \\
    \midrule
    \multirow{3}{*}{H20} & FastDeploy & 743.7 & 1.8206 & 2545.0 & 77.2 \\ & vLLM & 796.1 & 1.7007 & 2382.0 & 75.0 \\
                          & SGLang & 862.2 & 1.5702 & 2204.5 & 74.3 \\
    \midrule
    \multirow{3}{*}{L20} & FastDeploy & 845.0 & 1.6023 & 2248.4 & 41.0 \\ & vLLM & 998.2 & 1.3565  & 1890.7 & 25.1 \\
                          & SGLang & 1126.8 & 1.2018 & 1680.7 & 30.2 \\
    \midrule
    \multirow{3}{*}{A10} & FastDeploy & 1179.9 & 1.1477 & 1607.8 & 21.8 \\ & vLLM & 1245.5 & 1.0873 & 1520.6 & 13.5 \\
                          & SGLang & 1504.3 & 0.9003 & 1260.1 & 19.0 \\
    \midrule
    \multirow{2}{*}{RTX 3060} & vLLM & 2531.8 & 0.5351 & 748.1 & 11.8 \\
                              & SGLang & 2587.7 & 0.5235 & 730.5 & 11.7 \\
    \midrule
    \multirow{2}{*}{RTX 4090D} & vLLM & 923.5 & 1.4667 & 2040.1 & 16.3 \\
                               & SGLang & 1079.5 & 1.2548 & 1750.9 & 20.0 \\
    \bottomrule
  \end{tabular}%
    \caption{End-to-End Inference Performance}
      \label{tab:inference_performance_hardware}
\end{table}

\clearpage 
\newpage

\section{Real-world Samples}

This appendix demonstrates the robustness and versatility of PaddleOCR-VL-1.5 in processing diverse, high-complexity real-world scenarios.

\begin{figure}[H]
\centering
\includegraphics[width=0.88\linewidth]{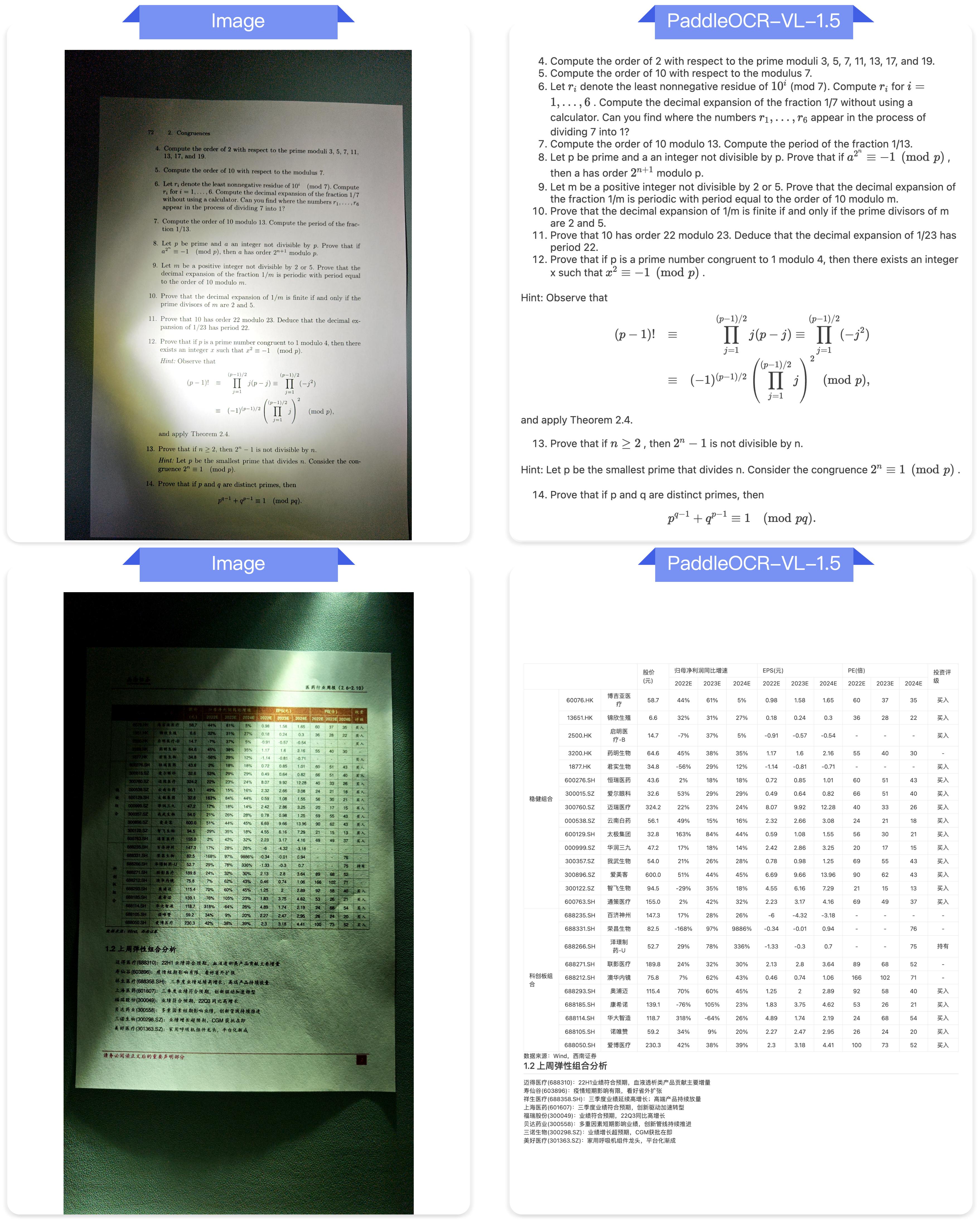} 

\caption{
    \centering
     The Markdown Output for Illumination.
}
\label{fig:lightall}
\end{figure}

\begin{figure}[H]
\centering
\includegraphics[width=0.95\linewidth]{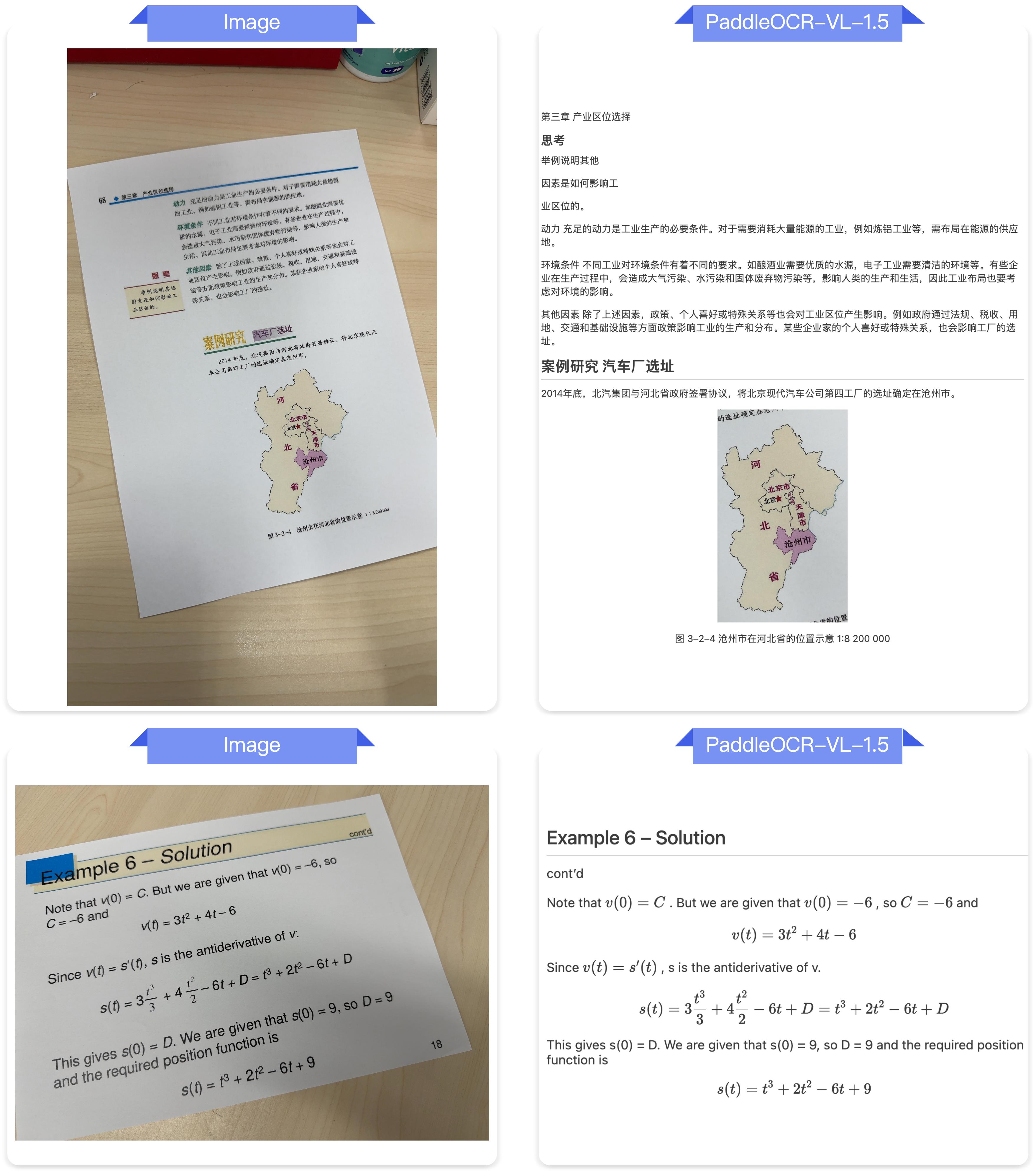} 

\caption{
    \centering
    The Markdown Output for Skew.
}
\label{fig:skewall}
\end{figure}

\begin{figure}[H]
\centering
\includegraphics[width=0.95\linewidth]{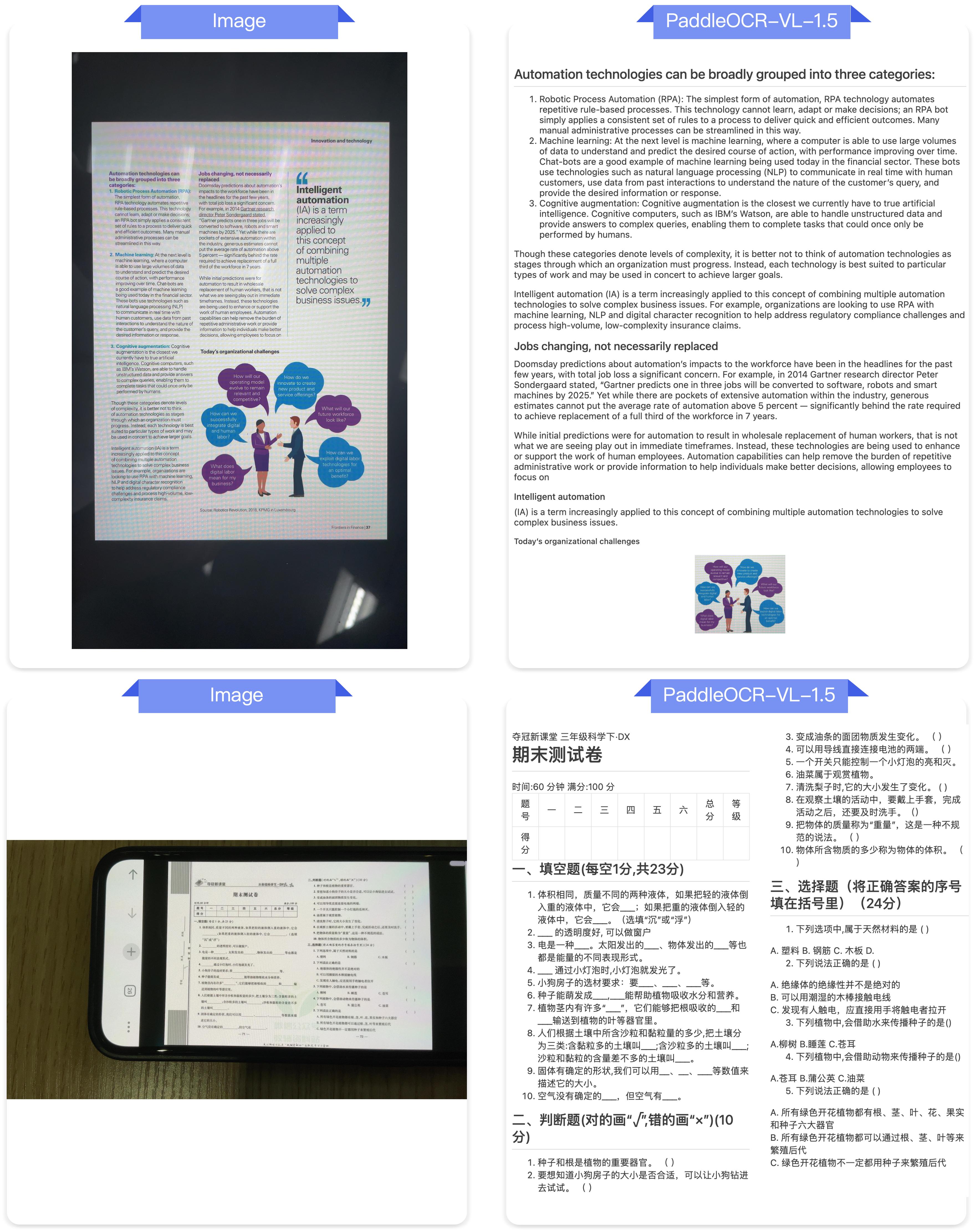} 

\caption{
    \centering
    The Markdown Output for Screen Photography.
}
\label{fig:screenall}
\end{figure}

\begin{figure}[H]
\centering
\includegraphics[width=0.95\linewidth]{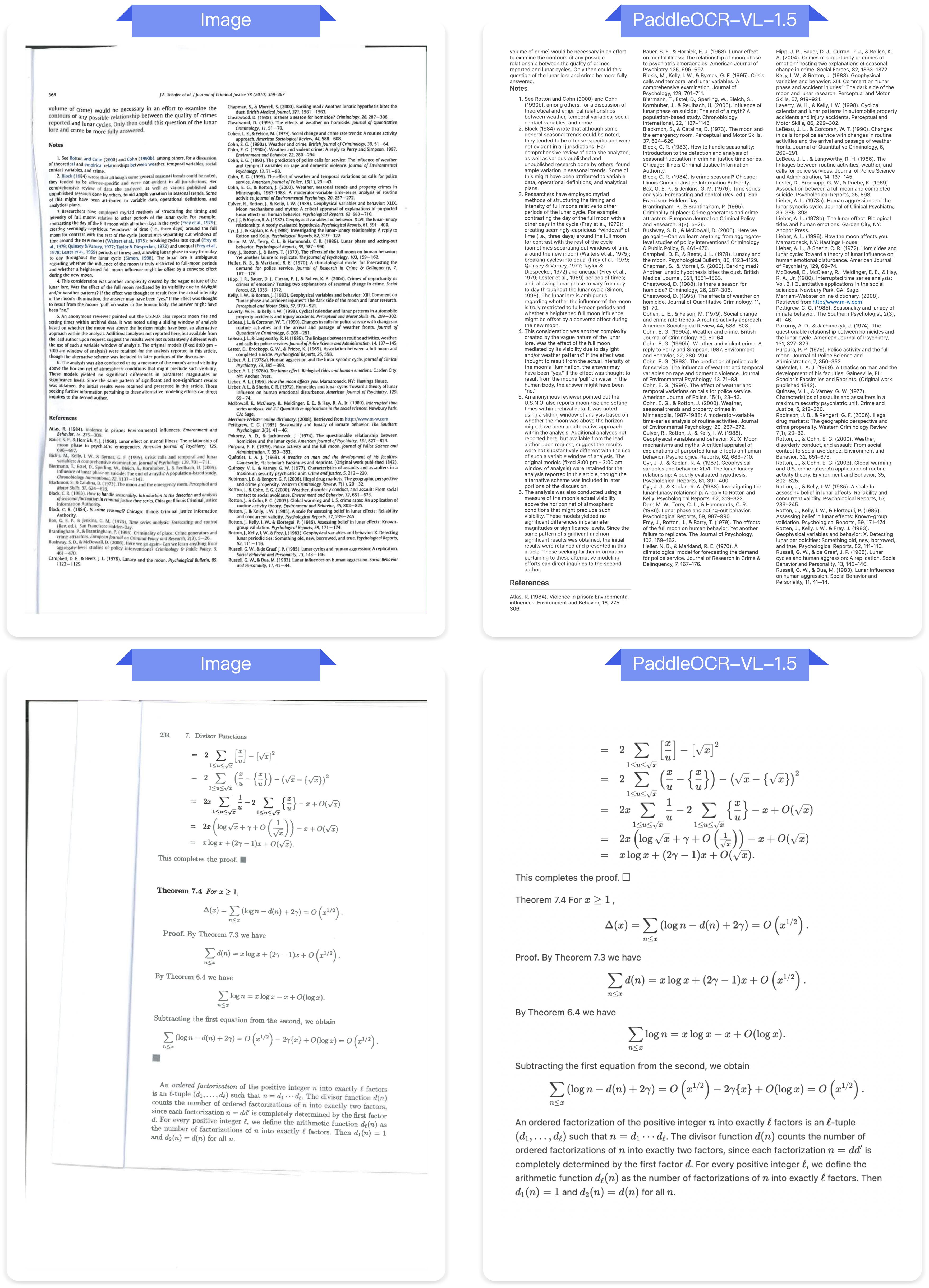} 

\caption{
    \centering
    The Markdown Output for Scanning.
}
\label{fig:scaning}
\end{figure}

\begin{figure}[H]
\centering
\includegraphics[width=0.95\linewidth]{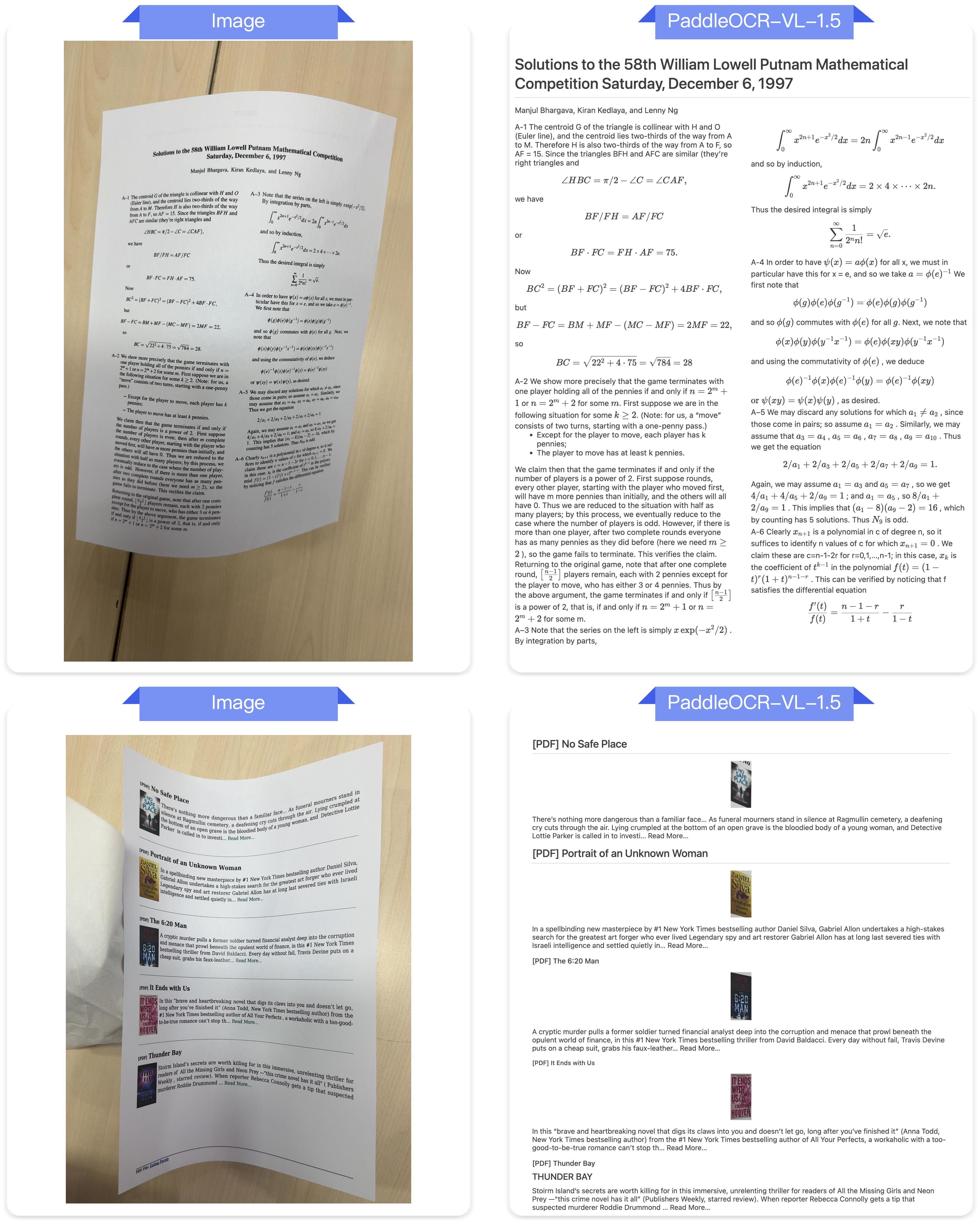} 

\caption{
    \centering
    The Markdown Output for Warping.
}
\label{fig:curvall}
\end{figure}

Section~\ref{subsec:Real-word Document Parsing} demonstrates the real-world document parsing capability of PaddleOCR-VL-1.5. Figures~\ref{fig:lightall}--\ref{fig:curvall} demonstrate the robust performance of PaddleOCR-VL-1.5 to parse real-world documents across diverse conditions, including varying illumination, geometric skew, screen-to-photo noise, and warped scanned surfaces.

\begin{figure}[H]
\centering
\includegraphics[width=0.90\linewidth]{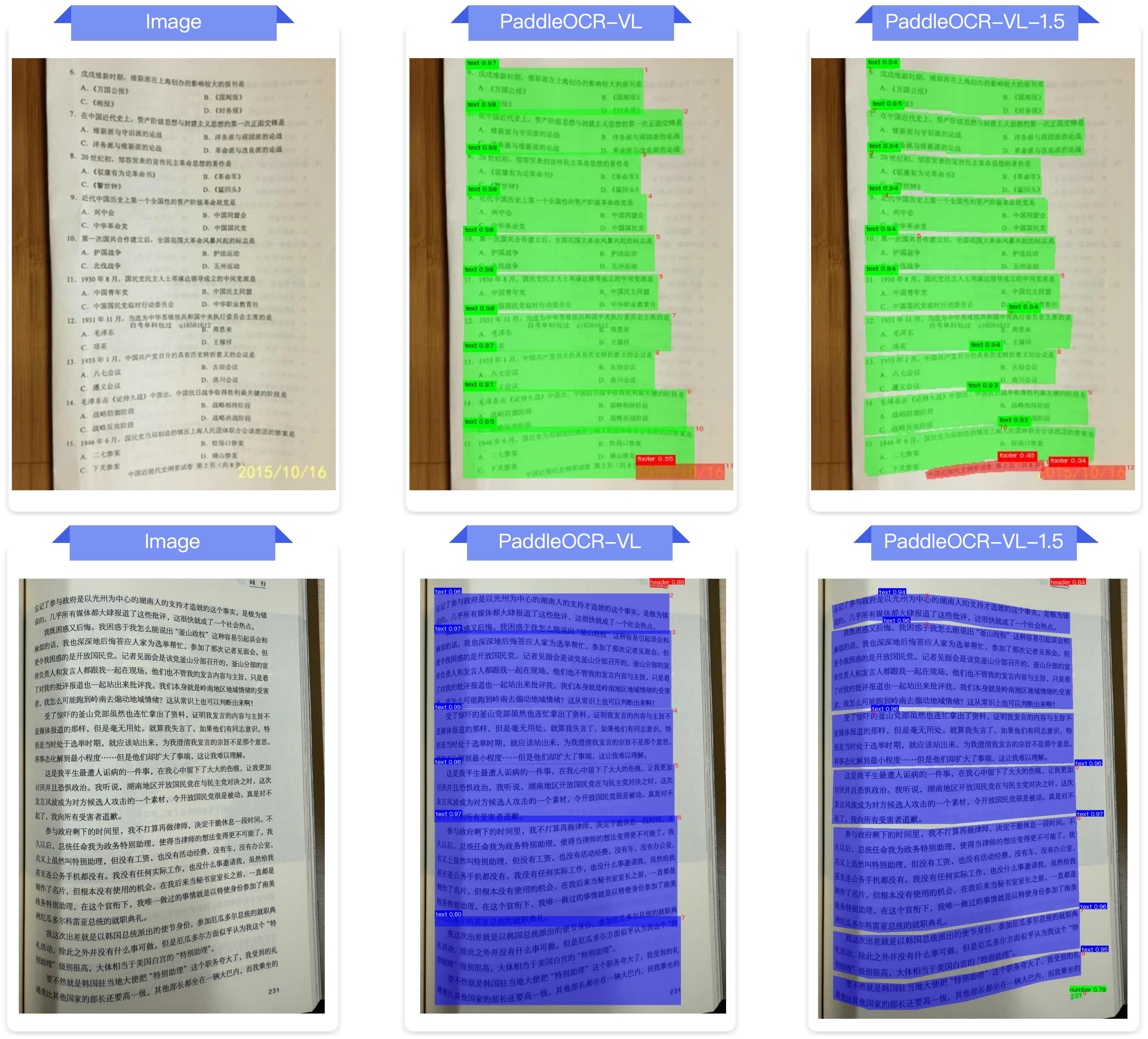} 

\caption{
    \centering
    Comparison of Layout Analysis Results between PaddleOCR-VL and PaddleOCR-VL-1.5 for Warping.
}
\label{fig:layout01}
\end{figure}

\begin{figure}[H]
\centering
\includegraphics[width=0.95\linewidth]{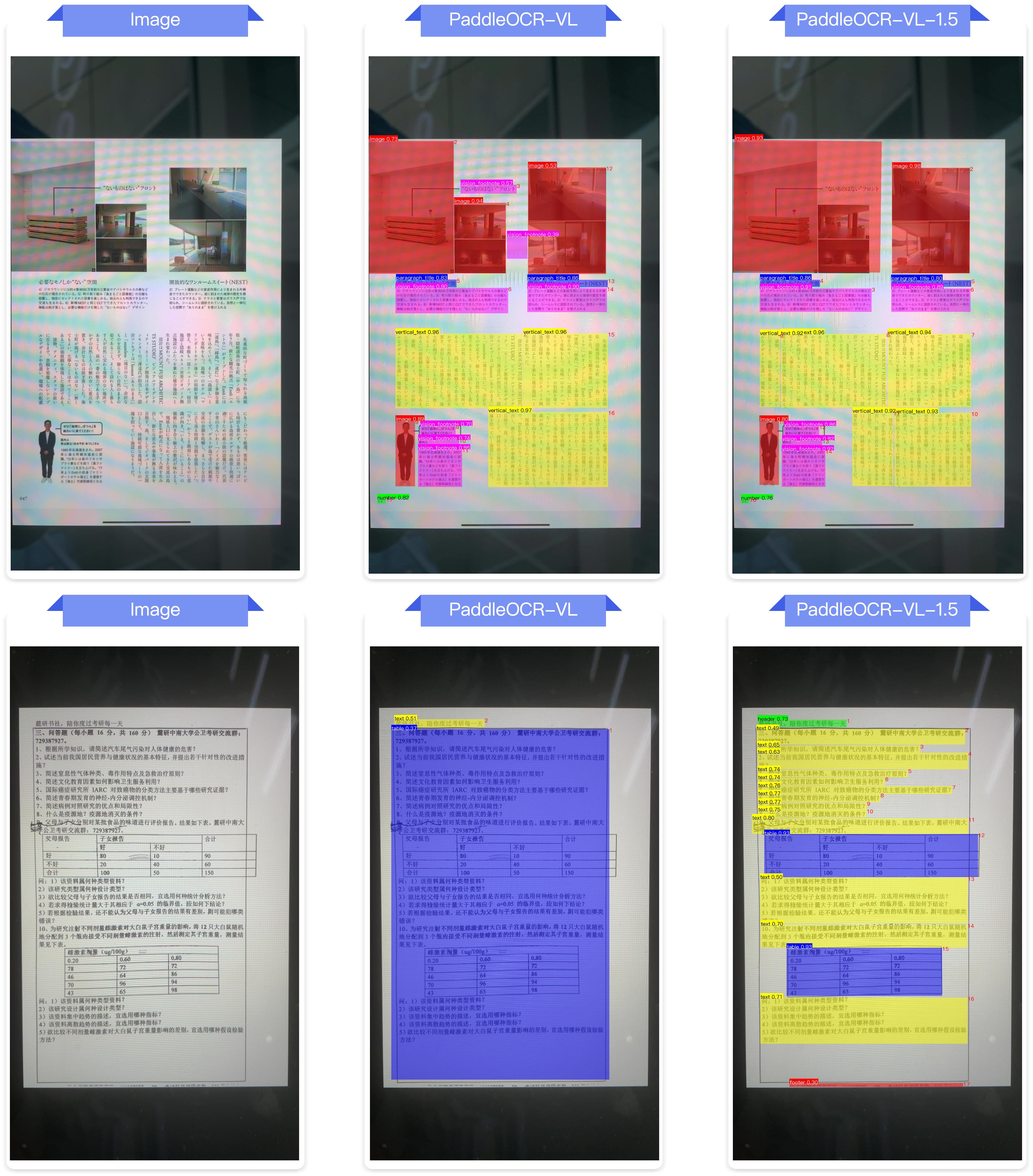} 

\caption{
    \centering
    Comparison of Layout Analysis Results between PaddleOCR-VL and PaddleOCR-VL-1.5 for Screen Photography.
}
\label{fig:layout02}
\end{figure}

\begin{figure}[H]
\centering
\includegraphics[width=0.95\linewidth]{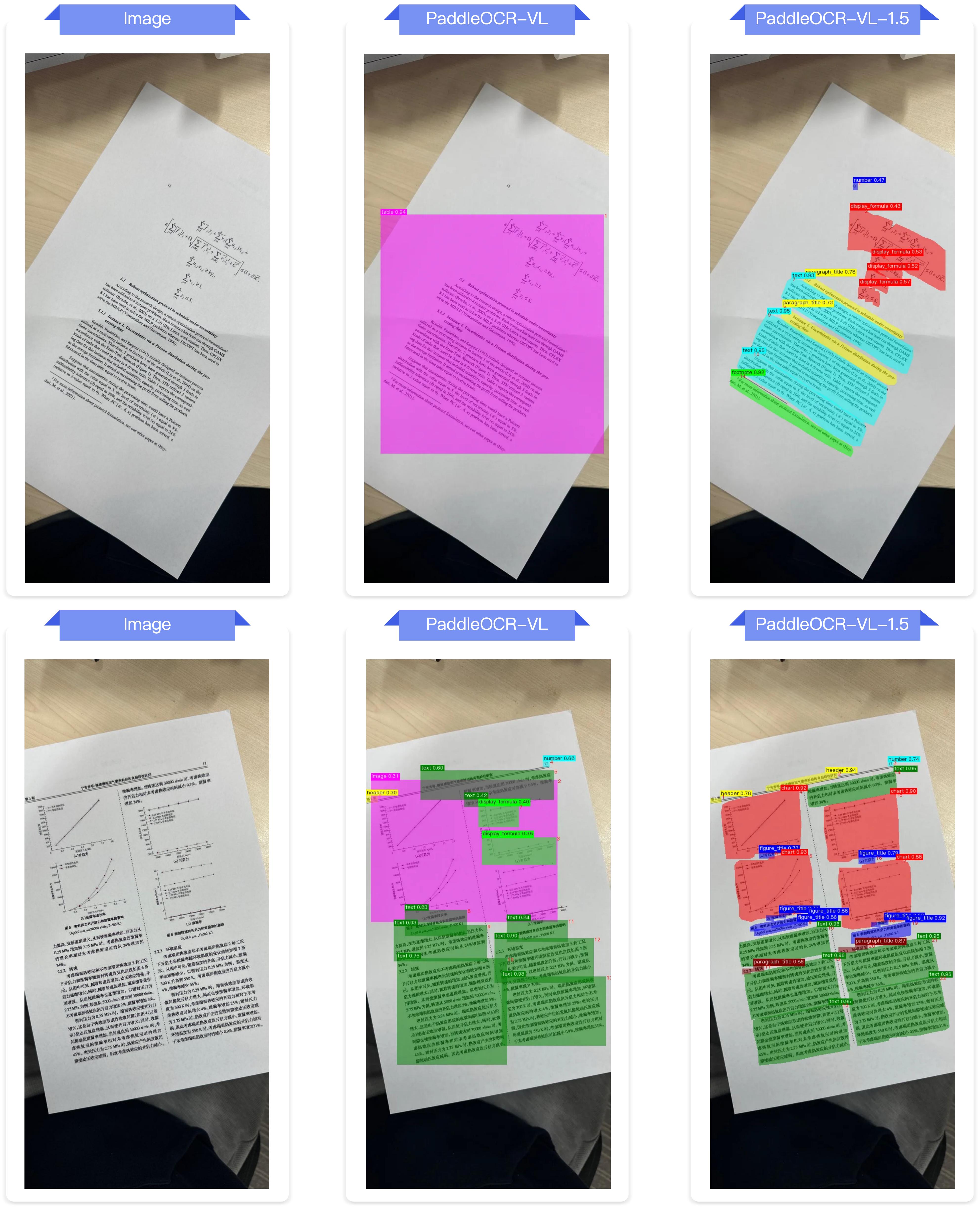} 

\caption{
    \centering
    Comparison of Layout Analysis Results between PaddleOCR-VL and PaddleOCR-VL-1.5 for Skew.
}
\label{fig:layout03}
\end{figure}

\begin{figure}[H]
\centering
\includegraphics[width=0.95\linewidth]{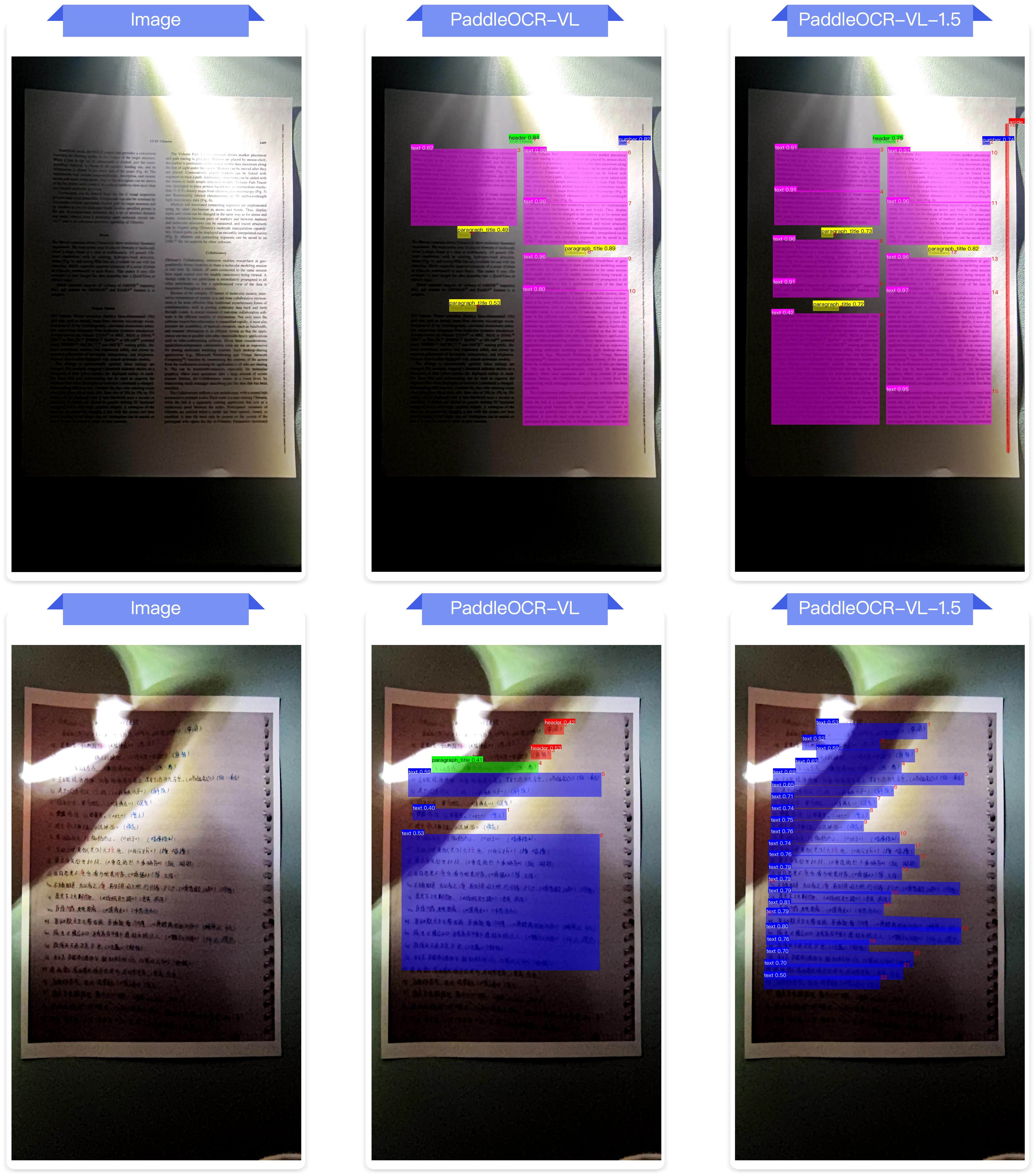} 

\caption{
    \centering
    Comparison of Layout Analysis Results between PaddleOCR-VL and PaddleOCR-VL-1.5 for Illumination.
}
\label{fig:layout04}
\end{figure}

\begin{figure}[H]
\centering
\includegraphics[width=0.92\linewidth]{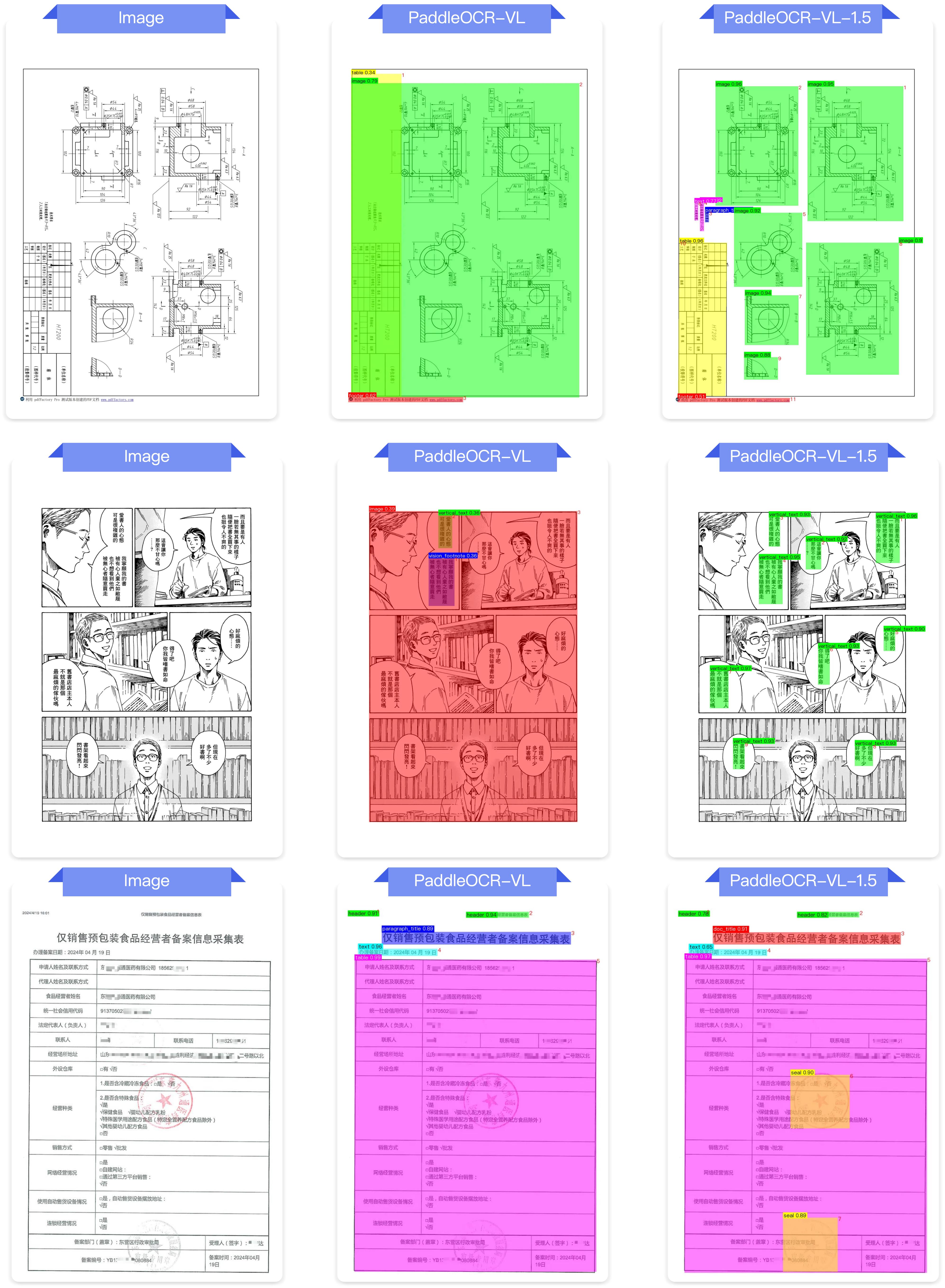} 

\caption{
    \centering
    Comparison of Layout Analysis Results between PaddleOCR-VL and PaddleOCR-VL-1.5 for New Scenarios.
}
\label{fig:layout05}
\end{figure}

Figures~\ref{fig:layout01}--\ref{fig:layout04} in Section~\ref{subsec:Layout Analysis} illustrate the robustness of PaddleOCR-VL-1.5 in layout analysis across challenging real-world conditions, including skewed or curved geometries, screen-to-photo noise, and illumination variations. Furthermore, Figure~\ref{fig:layout05} highlights its extended generalizability to specialized domains—such as comics, CAD drawings, and multi-stamped documents—where earlier version of the model faced limitations.

\begin{figure}[H]
\centering
\includegraphics[width=0.98\linewidth]{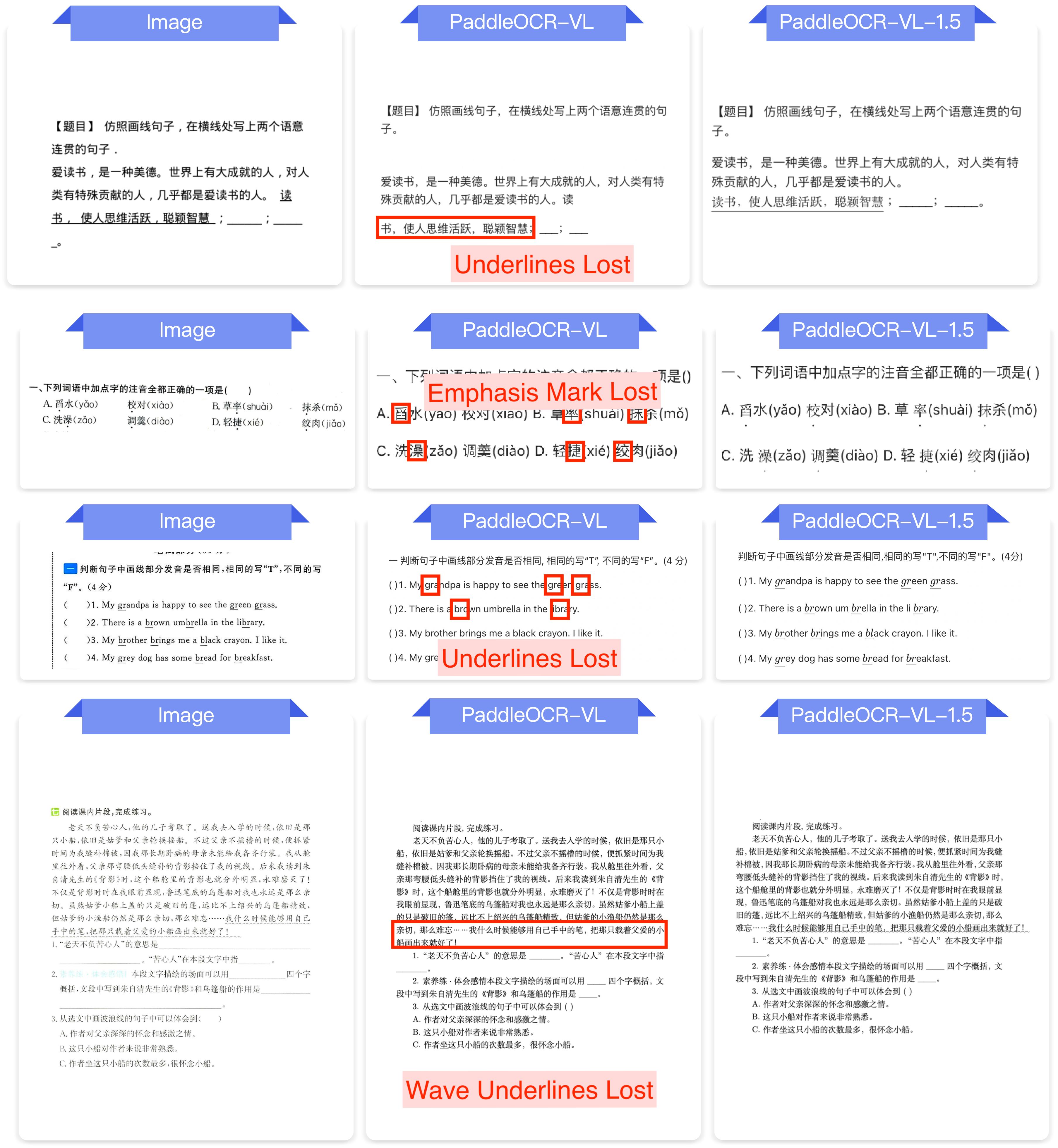} 

\caption{
    \centering
    Markdown Output Comparison between PaddleOCR-VL and PaddleOCR-VL-1.5 on Text Decoration Documents.
}
\label{fig:decoration}
\end{figure}

\begin{figure}[H]
\centering
\includegraphics[width=0.98\linewidth]{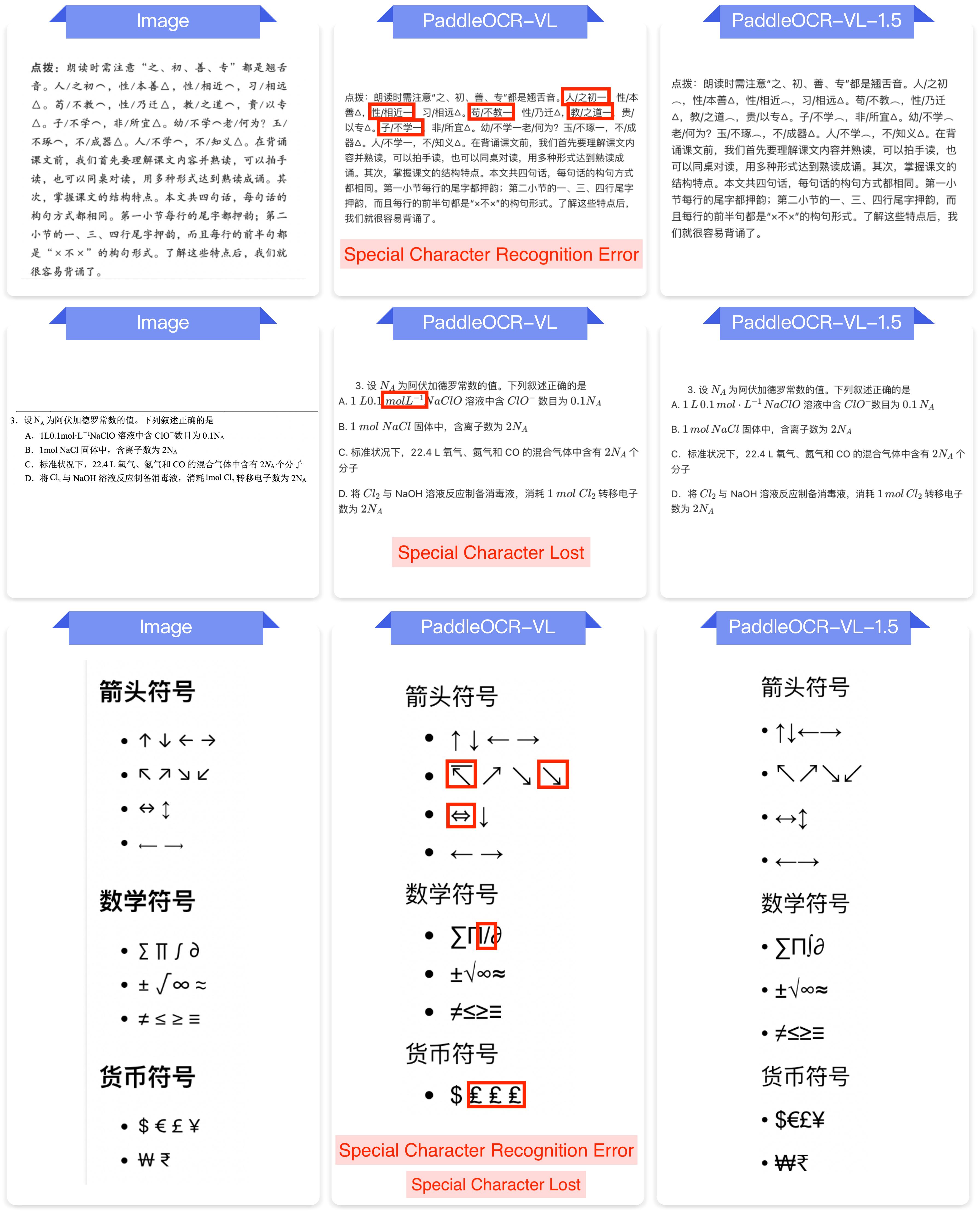} 

\caption{
    \centering
    Markdown Output Comparison between PaddleOCR-VL and PaddleOCR-VL-1.5 on Special Characters Documents.
}
\label{fig:special_characters}
\end{figure}

\begin{figure}[H]
\centering
\includegraphics[width=0.82\linewidth]{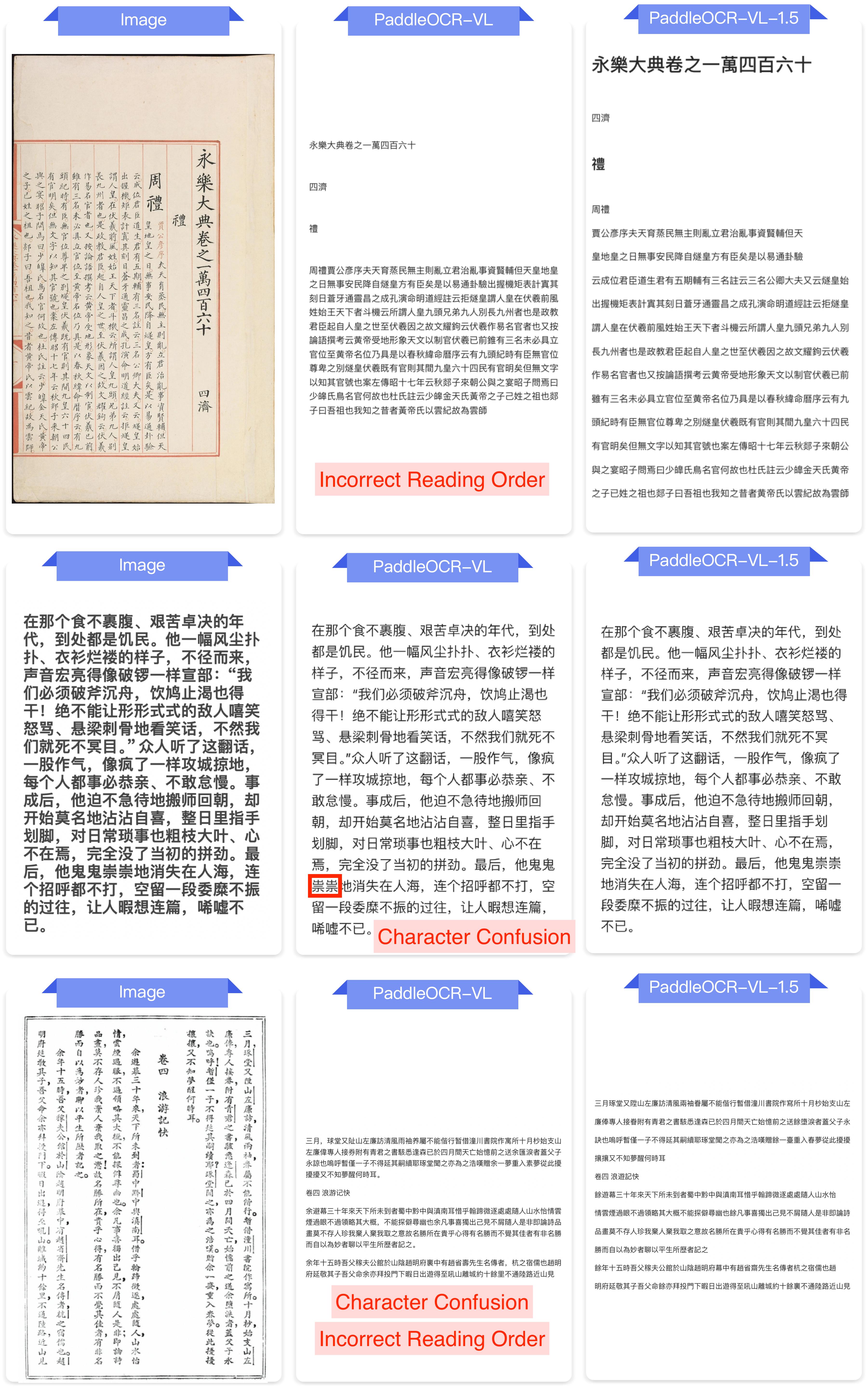} 

\caption{
    \centering
    Markdown Output Comparison between PaddleOCR-VL and PaddleOCR-VL-1.5 on Long-tail Scenarios Documents.
}
\label{fig:longtail}
\end{figure}

Section~\ref{subsec:Text Recognition} evaluates the text recognition performance of PaddleOCR-VL-1.5 under diverse constraints. As shown in Figure~\ref{fig:decoration}, the model exhibits improved sensitivity to text decorations, including underlines, emphasis marks, and wavy patterns, surpassing its predecessor. Figure~\ref{fig:special_characters} and Figure~\ref{fig:longtail} demonstrate enhanced robustness in identifying special characters and long-tail cases, such as vertical orientations and character-level ambiguities.

\begin{figure}[H]
\centering
\includegraphics[width=0.95\linewidth]{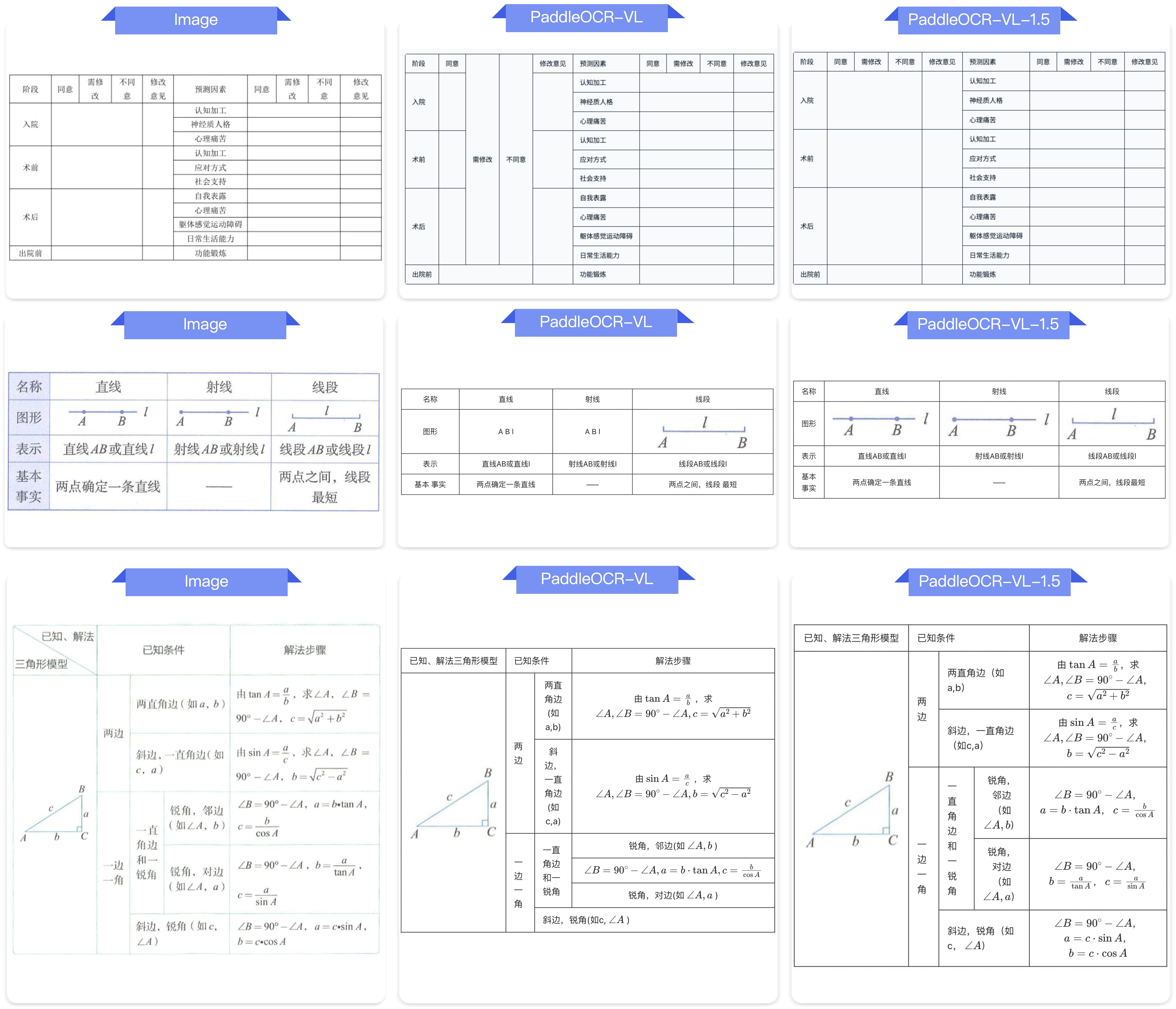} 

\caption{
    \centering
    Markdown Output Comparison between PaddleOCR-VL and PaddleOCR-VL-1.5 on General Tables.
}
\label{fig:table_01}
\end{figure}

\begin{figure}[H]
\centering
\includegraphics[width=0.95\linewidth]{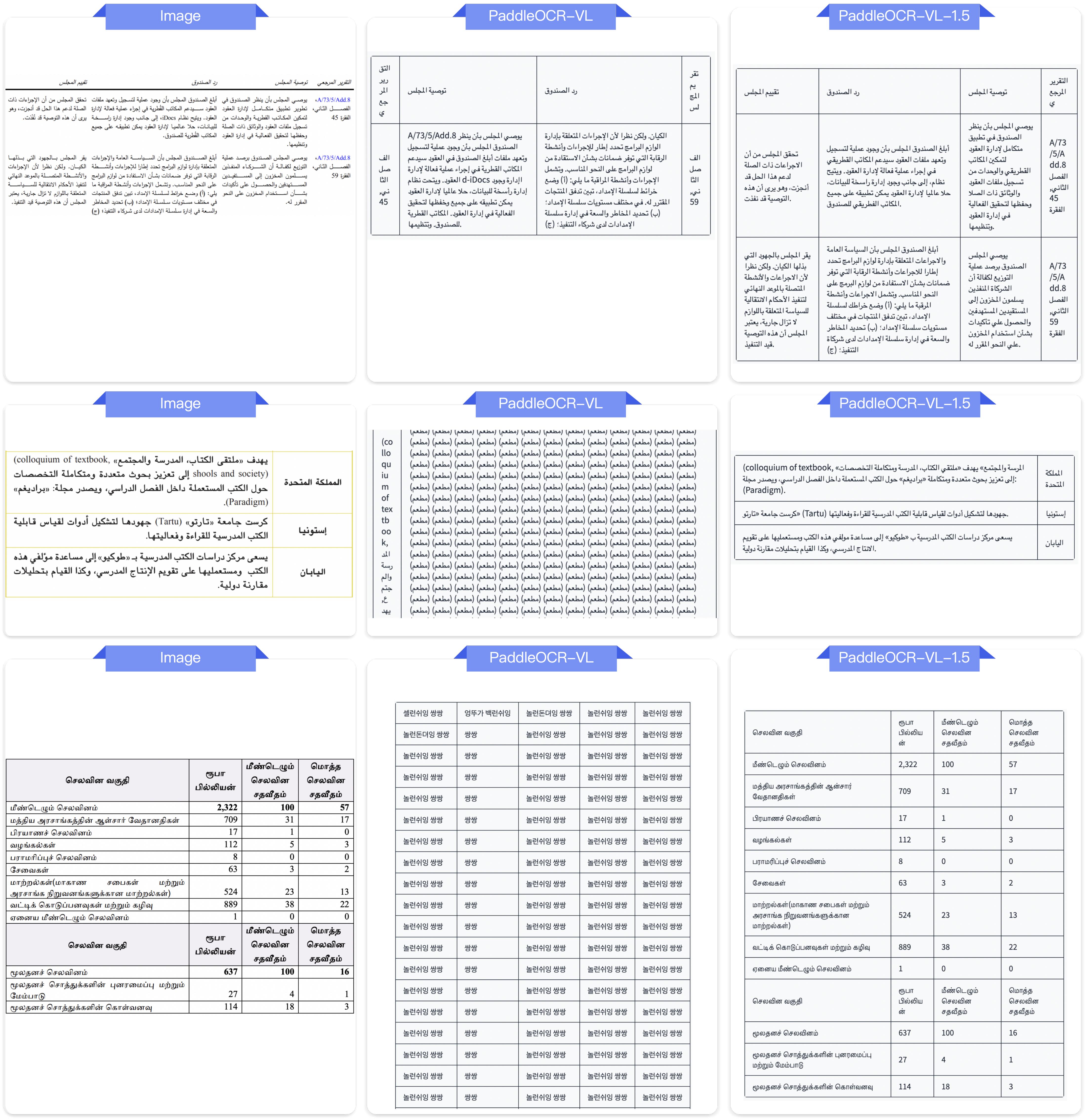} 

\caption{
    \centering
    Markdown Output Comparison between PaddleOCR-VL and PaddleOCR-VL-1.5 on Multilingual Tables.
}
\label{fig:table_02}
\end{figure}

\begin{figure}[H]
\centering
\includegraphics[width=0.95\linewidth]{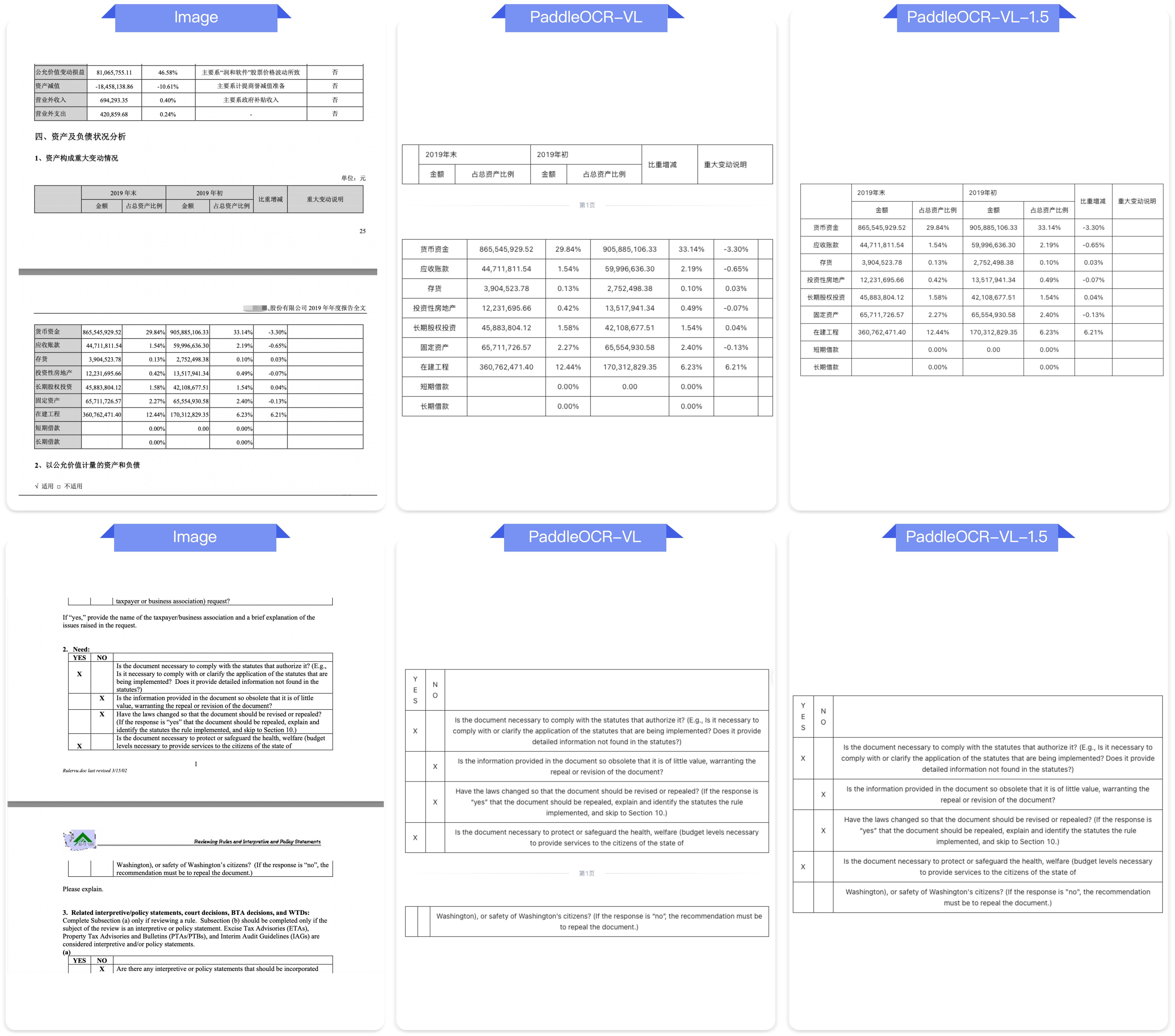} 

\caption{
    \centering
    Markdown Output Comparison between PaddleOCR-VL and PaddleOCR-VL-1.5 on Cross-Page Tables.
}
\label{fig:table_03}
\end{figure}

The model's table recognition abilities are demonstrated in section~\ref{subsec:Table Recognition}. Figure~\ref{fig:table_01} illustrates the robustness of the model in processing complex layouts, including tables from academic textbooks and those containing embedded images or mathematical formulas. The proficiency of the model in multilingual table recognition is presented in Figure~\ref{fig:table_02}. Furthermore, Figure~\ref{fig:table_03} demonstrates its extended capability for cross-page table detection and merging, addressing the challenges of multi-page document parsing."

\begin{figure}[H]
\centering
\includegraphics[width=0.95\linewidth]{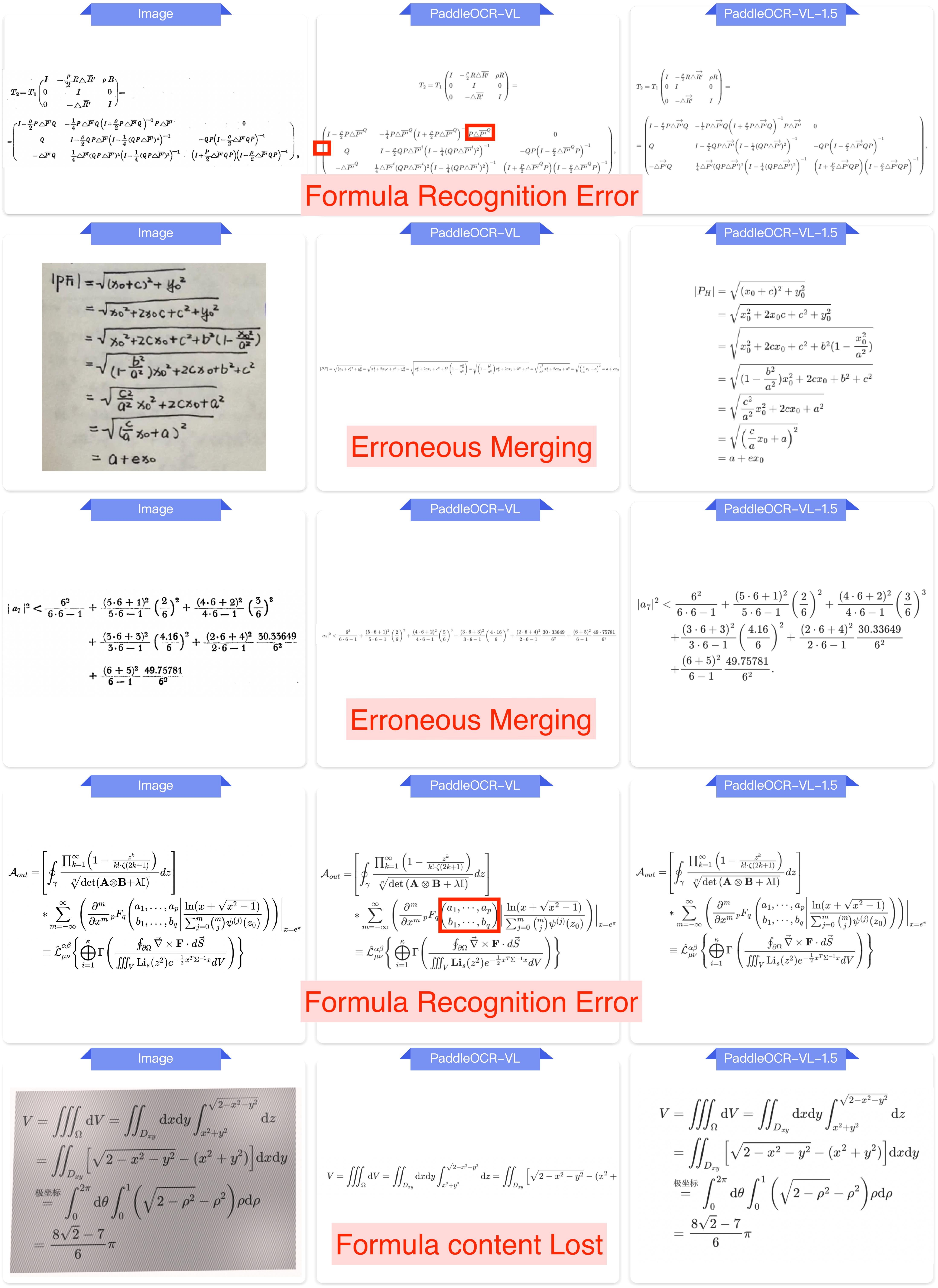} 

\caption{
    \centering
    Markdown Output Comparison between PaddleOCR-VL and PaddleOCR-VL-1.5 on various types of Formulas.
}
\label{fig:formula1}
\end{figure}

Figures in section~\ref{subsec:Formula Recognition} detail the formula recognition performance. As illustrated in Figure \ref{fig:formula1} , the updated model exhibits superior performance in mathematical expression recognition, specifically regarding sub/superscript accuracy, multi-line formula segmentation, and a lower overall error rate.

\begin{figure}[H]
\centering
\includegraphics[width=0.80\linewidth]{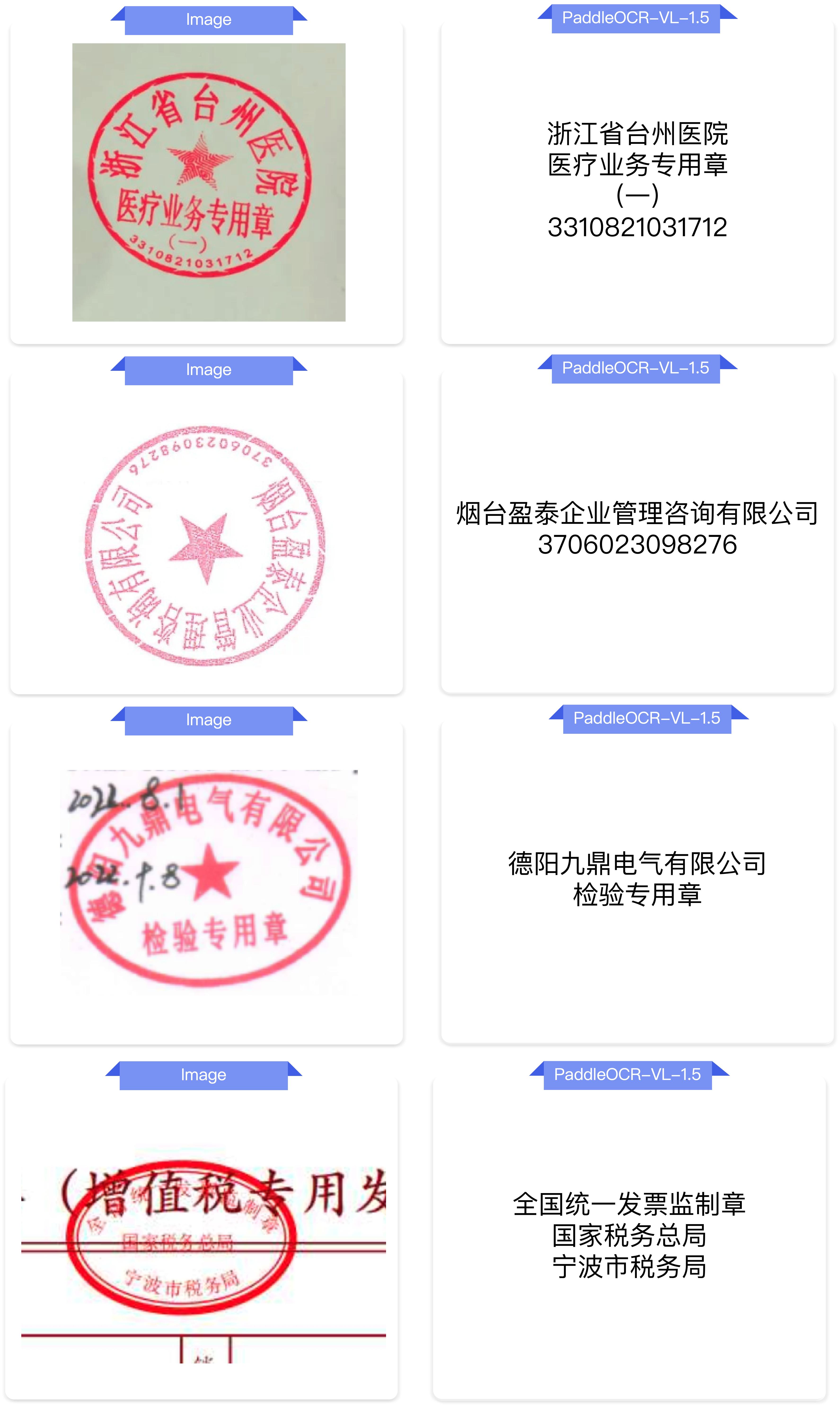} 

\caption{
    \centering
    The markdown output for various types of Seals 1.
}
\label{fig:seal1}
\end{figure}

\begin{figure}[H]
\centering
\includegraphics[width=0.80\linewidth]{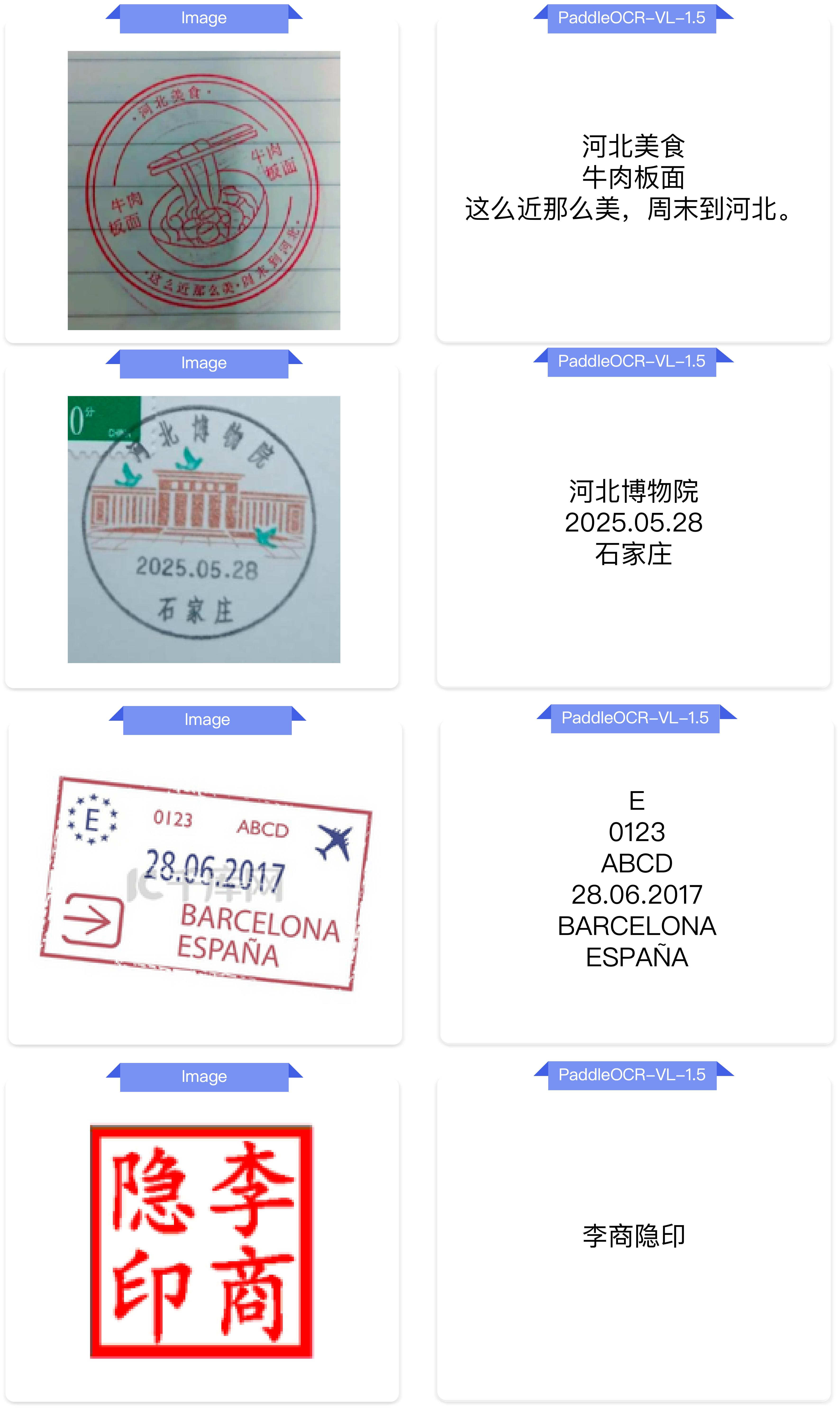} 

\caption{
    \centering
    The markdown output for various types of Seals 2.
}
\label{fig:seal2}
\end{figure}

\begin{figure}[H]
\centering
\includegraphics[width=0.80\linewidth]{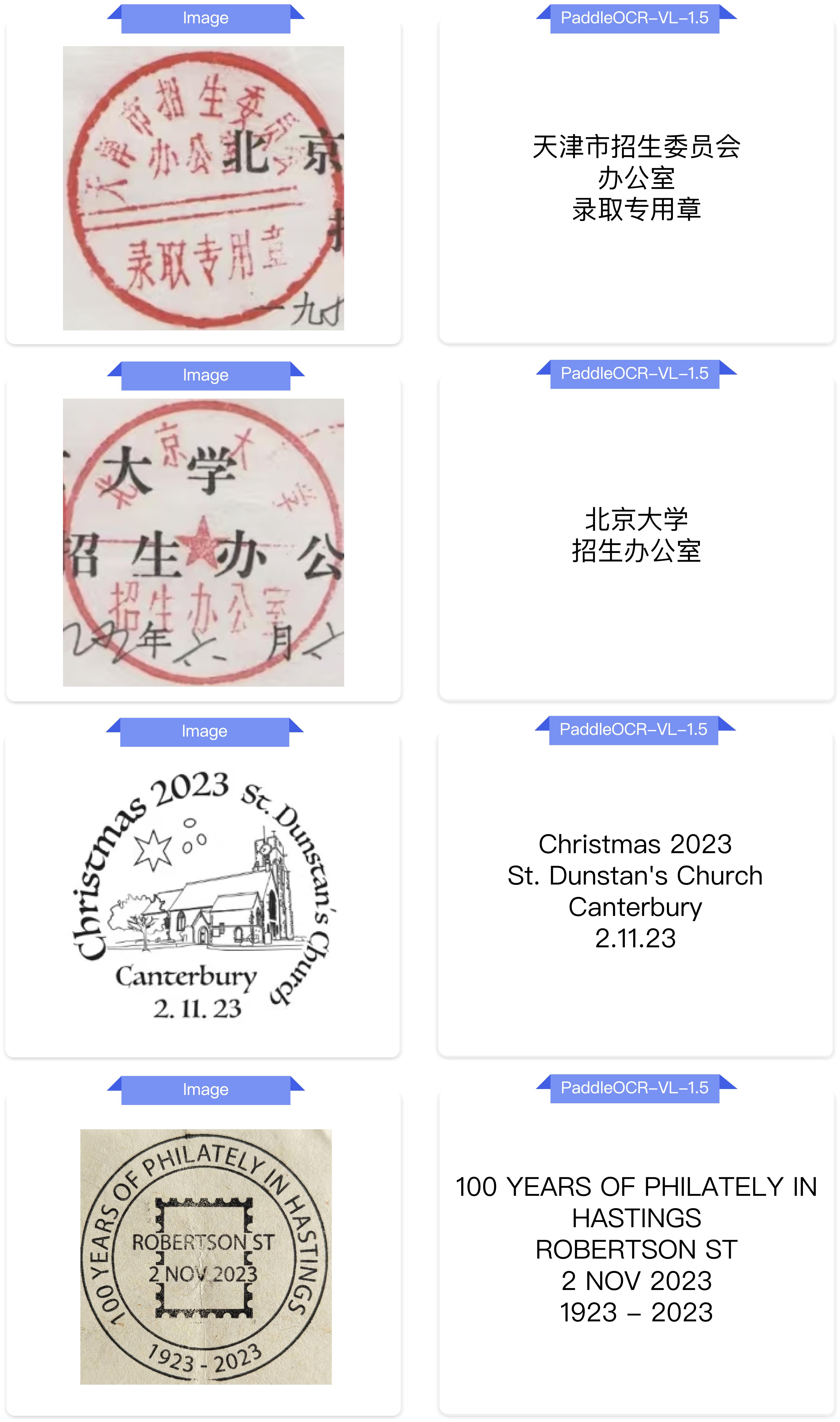} 

\caption{
    \centering
    The markdown output for various types of Seals 3.
}
\label{fig:seal3}
\end{figure}

In section~\ref{subsec:Seal Recognition}, seal recognition represents a novel capability of this model update. As illustrated in Figures~\ref{fig:seal1}--\ref{fig:seal3}, the model accurately extracts content from various types of seals, demonstrating high precision even when faced with complex background interference and cluttered environments.

\begin{figure}[H]
\centering
\includegraphics[width=0.95\linewidth]{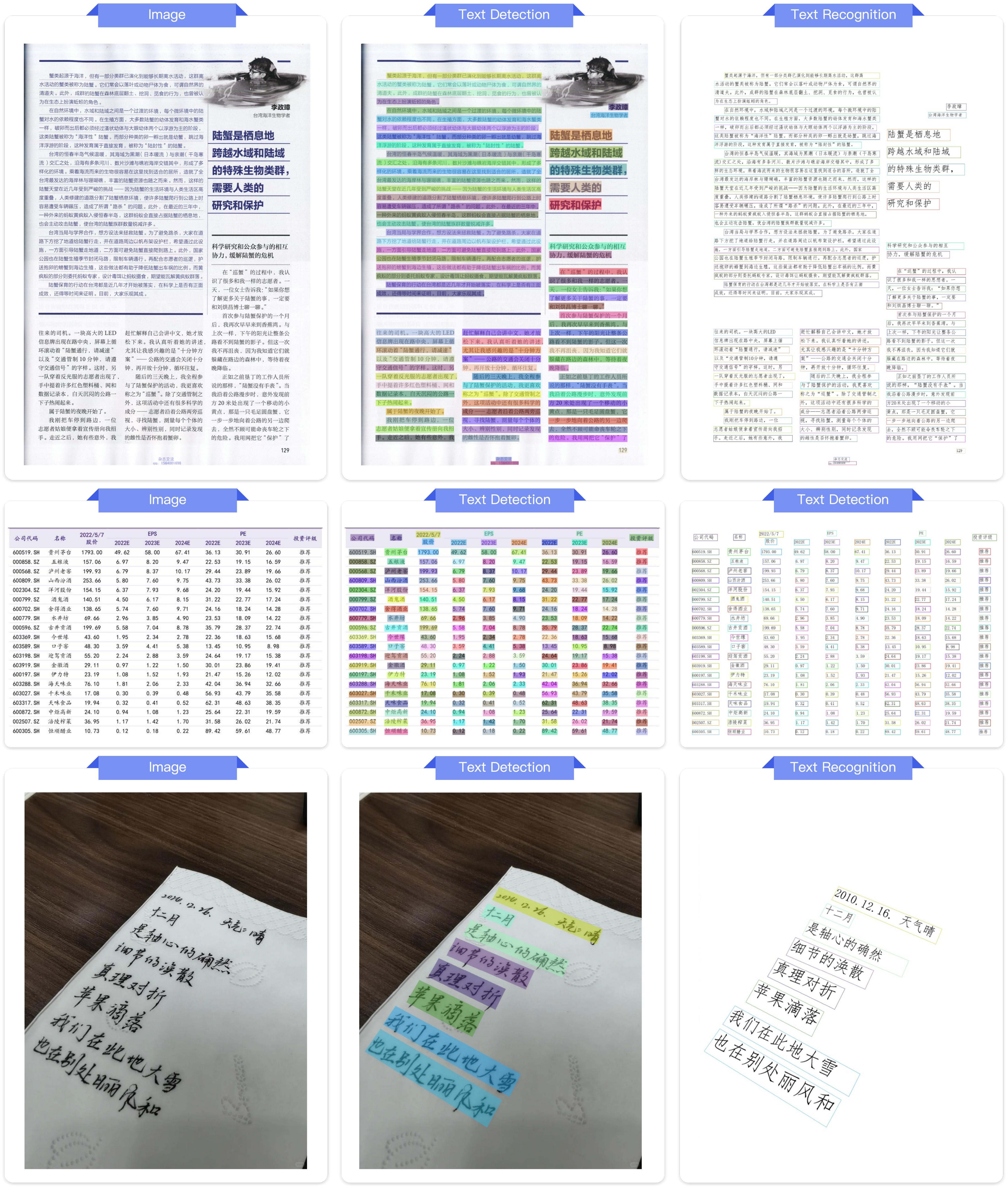} 

\caption{
    \centering
    Text spotting results on various types of documents.
}
\label{fig:spotting}
\end{figure}

Figure~\ref{fig:spotting} in section~\ref{subsec:Text Spotting} highlights the newly integrated text spotting capability of the model, which enables simultaneous localization and recognition. The results demonstrate superior robustness across challenging layouts, ranging from multi-column magazine pages and complex tables to irregular handwritten content.

\clearpage 
\newpage
\subsection{Real-word Document Parsing}
\label{subsec:Real-word Document Parsing}

\clearpage 
\newpage

\clearpage 
\newpage

\clearpage 
\newpage

\clearpage 
\newpage




\newpage
\subsection{Layout Analysis}
\label{subsec:Layout Analysis}

\subsubsection{Layout Analysis for Real-world Documents}
\label{subsec:Layout Analysis for Real-world Documents}

\newpage
\subsubsection{Layout Analysis for New Scenarios}
\label{subsec:Layout Analysis in New Scenarios}

\newpage
\subsection{Text Recognition}
\label{subsec:Text Recognition}

\subsubsection{Text Recognition for Text decoration}
\label{subsec:Text Recognition for Text decoration}

\subsubsection{Text Recognition for Special characters}

\clearpage 
\newpage
\subsubsection{Text Recognition for Long-tail Scenarios}

\clearpage 
\newpage

\subsection{Table Recognition}
\label{subsec:Table Recognition}

\subsubsection{Table Recognition for General Tables}
\label{subsec:Table Recognition for Multiple Languages}

\subsubsection{Table Recognition for Multiple Languages}
\label{subsec:Table Recognition for General Tables}

\subsubsection{Table Recognition for Cross-Page Tables}
\label{subsec:Table Recognition for Cross-Page Tables}

\clearpage 
\newpage

\subsection{Formula Recognition}
\label{subsec:Formula Recognition}

\clearpage 
\newpage

\subsection{Seal Recognition}
\label{subsec:Seal Recognition}

\clearpage 
\newpage

\subsection{Text Spotting}
\label{subsec:Text Spotting}

\end{document}

%% file: commands.tex
\renewcommand{\phi}{\varphi}

\renewcommand{\epsilon}{\varepsilon}
\renewcommand{\imath}{\mathrm{i}}

\newlength{\restsubwidth}
\newlength{\restsubheight}
\newlength{\restsubmoreheight}
\newcommand{\rest}[2]{%
        \settowidth{\restsubwidth}{\ensuremath{#2}}
        \settoheight{\restsubheight}{\ensuremath{{}_{#2}}}
        \ensuremath{{#1\hskip 0.5pt}_{\vrule\kern2pt\parbox[b][%
        4pt][b]{\the\restsubwidth}{%
                        \ensuremath{{}_{#2}}}}}
        }